\documentclass[11pt]{article}

\usepackage{acl}

\usepackage{times}
\usepackage{latexsym}
\usepackage[T1]{fontenc}
\usepackage[utf8]{inputenc}
\usepackage{microtype}
\usepackage{inconsolata}
\usepackage{graphicx}
\usepackage{amsmath}
\usepackage{amssymb}
\usepackage{booktabs}
\usepackage{algorithm}
\usepackage{algpseudocode}
\usepackage{url}
\usepackage{booktabs}
\usepackage{multirow}
\usepackage[table]{xcolor}
\usepackage{tcolorbox}
\tcbuselibrary{skins} 
\usepackage{enumitem} 
\usepackage{subcaption}
\usepackage{fontawesome5} 
\usepackage{pdfpages} 

\title{ViCo: Visual-oriented Coding with Self-Reflection for Chart Replication}

\author{
  \textbf{Jiaxin Duan},
  \textbf{Dian Jiao}
  \textbf{Shuai Zhao},
  \textbf{Jiabing Leng},
  \textbf{Yiran Zhang},
  \textbf{Feng Huang} \\  %
  China Electronics Cloud Technology Co., Ltd. \\
  \texttt{\{duanjiaxin,jiaodian,zhaoshuai\}@cestc.cn} \\
  \texttt{\{lengjiabing,zhangyiran,huangfeng01\}@cestc.cn} \\
  \small\faGithub\ \url{https://github.com/Xigmoid/ViCo}
}

\begin{document}
\maketitle

\renewcommand{\thefootnote}{\fnsymbol{footnote}} 



\begin{abstract}
This paper addresses the challenge of generating high-quality academic charts that match the visual standards of human-authored papers. While existing AI agents can produce well-structured text and code, their generated visualizations often lack the stylistic and semantic fidelity of human designs. Advanced coding agents that employ self-reflection mechanisms exhibit poor visual reasoning and limited reflection following, resulting in sparse reward signals that severely undermine their reinforcement learning (RL). We propose ViCo, a training framework for visual-oriented coding that employs iterative reflections to align generated chart images progressively with the reference. We first introduce a self-supervised warm-up stage, which augments \textit{Monte Carlo Tree Search} with consistency-based pruning to synthesize high-quality reflection trajectories, ensuring that each coding step strictly follows the outcomes of prior reflections. A multi-step RL algorithm is then developed, using \textit{counterfactual baselines} to estimate advantage for reflection and action steps within each refinement cycle, thereby addressing the reward sparsity. To enable efficient reward in massive training, we propose an automatic, multifaceted evaluation framework that assesses charts' style, layout, and semantic consistency via a hierarchical heterogeneous layout graph structure. Experiments on three public benchmarks demonstrate that ViCo, trained on an 8B model, achieves performance close to proprietary LLMs with adequate reflection capabilities.
\end{abstract}

\section{Introduction}

Scientific plotting is fundamental for academic publications, providing a concise visual representation of complex data that enables readers to grasp key findings, trends, and comparisons quickly. Despite significant advances in AI-powered academic assistants that can generate well-structured text and syntactically correct code~\cite{AutoResearchClaw,SciAgents}, the visual quality of automatically produced scientific plots still lags considerably behind those created by human researchers \cite{ChartCoder,TinyChart}. A promising direction to bridge this gap is to leverage retrieval-based methods to obtain high-quality reference images from the existing literature, and then guide an AI agent to replicate the visual style of these references while making only minimal adjustments to accommodate the actual data~\cite{zhu2026paperbanana,zhu2026autofigure}. This paradigm, referred to as chart-to-code generation~\cite{ChartCoder,jiang2025viscodexunifiedmultimodalcode}, has recently attracted growing attention due to its potential in producing lossless representations that preserve all critical details of a chart, including colors, coordinates, and data values.

Training an AI agent to faithfully reproduce reference plots from real data is challenging. On the one hand, the agent must possess \textbf{strong coding capabilities} to invoke appropriate toolkits (e.g., Matplotlib, Seaborn) and generate executable scripts that produce the desired visual output~\cite{RealChart2Code}. More critically, the robust visual reasoning capacity is required to compare its own rendered output against the reference image, identify discrepancies, and iteratively refine the code accordingly. To this end, several recent works have explored self-reflection mechanisms~\cite{Renze_2024,ji-etal-2023-towards}, where an agent alternates between code generation and self-critique to progressively improve its output. VisRefiner~\cite{VisRefiner} employs a reinforcement learning (RL) stage for self-refinement by observing visual differences between the rendered prediction and the target design. ChartSketcher~\cite{ChartSketcher} introduces a multimodal feedback-driven step-by-step reasoning method that annotates intermediate reasoning steps directly onto the chart.

Despite these advances, existing reflection-based coding agents still suffer from critical limitations. First, \textbf{the effectiveness of reflection is often questionable.} Empirical observations reveal that in many cases the agent fails to identify the root cause of the visual discrepancy, producing superficial or even misleading critiques that lead to negligible improvements or, worse, degrade the code quality over multiple iterations~\cite{ChartHal}. Second, the agent also exhibits \textbf{limited reflection-following capability.} Even when the reflection correctly identifies a flaw, the subsequent code generation step may ignore or misinterpret the suggested fix, resulting in a “say one thing, do another” behavior. Third, optimizing such multi-turn ReAct~\cite{ReAct} processes via RL faces the severe credit assignment challenge caused by reward sparsity. In a typical reflection loop, the global reward (e.g., replication fidelity) is only available at the end of a long trajectory. RL algorithms, like PPO~\cite{PPO} or GRPO~\cite{GRPO}, struggle to attribute this sparse global reward to specific coding or reflection steps within the trajectory~\cite{VisRefiner}. The lack of fine-grained credit assignment hinders the model's ability to learn constructive reflection and robust coding that follow its own introspective comments, leading to suboptimal convergence or even performance degradation during training. Additionally, existing approaches often rely on heavy proprietary models (e.g., GPT-5~\cite{singh2026openaigpt5card}) as judges for quality assessment, which is computationally prohibitive for large-scale training~\cite{ChartCoder}.

To address these challenges, we propose \textbf{ViCo}, a \textbf{Vi}sual-oriented \textbf{Co}ding framework that enhances chart replication through reflective self-training. To prevent unproductive reflections and execution failures, ViCo introduces a \textbf{self-supervised warm-up} phase using trajectories sampled via Monte Carlo Tree Search (MCTS) with \textbf{consistency-based pruning}. A \textit{checklist mechanism} is employed during MCTS simulation to evaluate whether the generated code strictly aligns with the preceding reflection. Trajectory paths that violate this consistency or fail to yield monotonic reward improvements are pruned immediately. This process produces high-quality trajectories in which each coding step demonstrably executes its corresponding reflection to achieve higher rewards, enabling effective behavior cloning of the self-reflection policy.
Subsequently, a counterfactual-based advantage estimating mechanism is introduced in multi-step RL to address credit assignment. By curating distinct counterfactual baselines for the reflection and coding steps within each reflection episode, ViCo generates dense, step-level learning signals, which effectively stabilizes RL training without requiring expensive process reward models. Finally, to scale ViCo on large datasets, we propose a lightweight, graph-grounded evaluation framework based on a Hierarchical Heterogeneous Layout Graph (HHLG). By modeling charts as hierarchical tree-graph structures, HHLG facilitates precise element alignment and assesses replication fidelity across key complementary dimensions, including semantic, layout, and style consistency, against ground-truth references. Empirical results ($\S$~\ref{sec:ablation}) reveal that our HHLG-based evaluation correlates strongly with human judgments, while remaining explainable and computationally efficient for joint warm-up and RL training.


Our contributions are three-folds:
1) We propose \textbf{ViCo}, a training framework that integrates self-supervised warm-up with counterfactual-based reinforcement learning, enabling robust and effective self-reflection in visual-oriented code generation for academic chart replication.
2) We design a lightweight yet structurally comprehensive reward framework based on a hierarchical heterogeneous layout graph, which provides stable, human-aligned evaluation signals and supports scalable reinforcement learning.
3) Extensive evaluations across three challenging benchmarks demonstrate that ViCo systematically improves code pass rates (exceeding 95\%) while yielding LLM-judged scores highly competitive with state-of-the-art proprietary closed-source models.


\begin{figure}[!!t]
\centering
\includegraphics[width=1.0\linewidth]{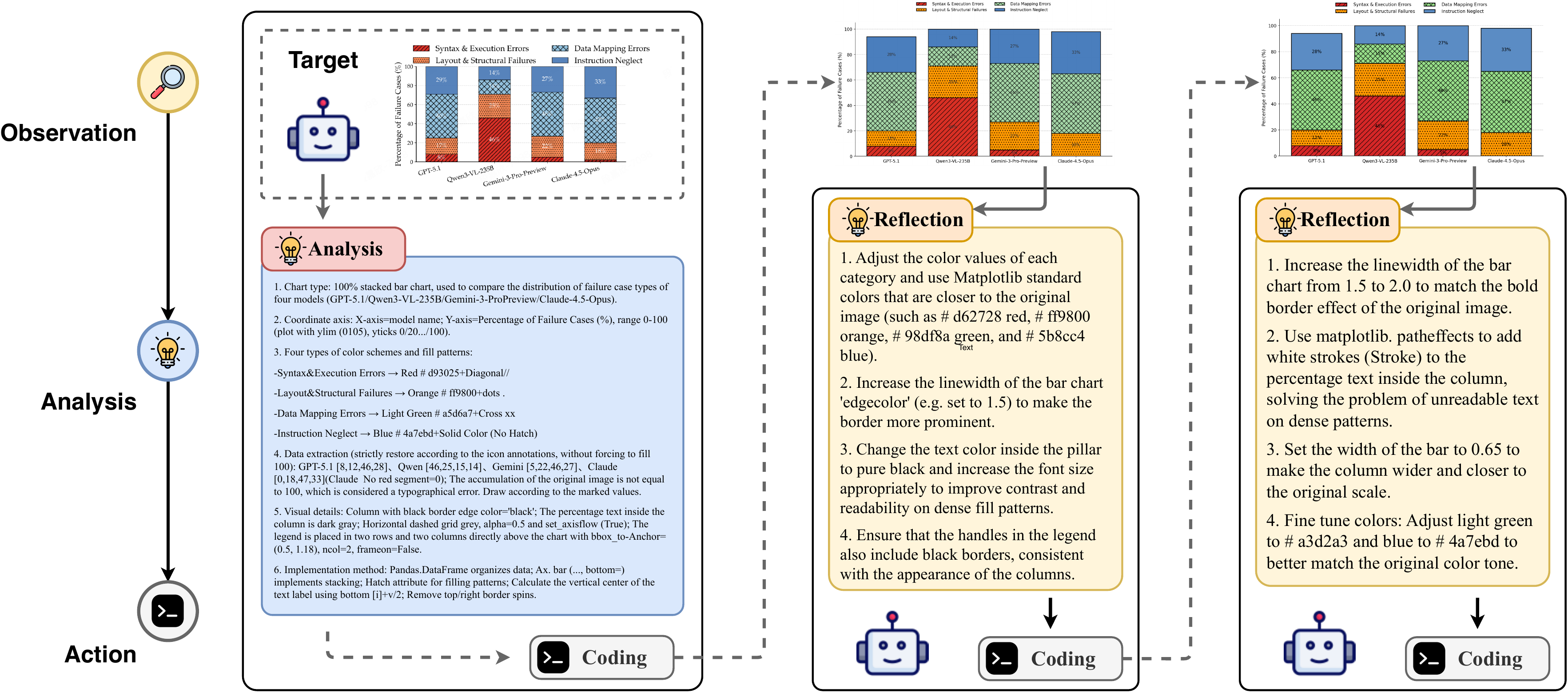}
\caption{Overview of visual coding with self-reflection.}
\label{fig:teaser}
\end{figure}

\section{Problem Formulation}
\label{sec:2.1}
ViCo replicates an academic chart $\mathcal{X}$ by alternating reflection and coding over $K$ consecutive steps, as shown in Figure~\ref{fig:teaser}. Starting from an initial code $a_0$ generated directly from $\mathcal{X}$, at each macro episode $t \in \{1, \dots, K\}$, the agent first performs a textual \textbf{reflection} action $r_t$ to analyze the visual discrepancies between the previous render and $\mathcal{X}$, and then executes a \textbf{coding} action $a_t$ to refine the code script. This iterative generation process yields a macro trajectory:
\[
\tau = \big( a_0, (r_1, a_1), (r_2, a_2), \dots, (r_K, a_K) \big).
\]

Mathematically, this sequential decision-making process is a finite-horizon Partially Observable Markov Decision Process (POMDP), defined by the tuple $\mathcal{M} = \langle \mathcal{S}, \mathcal{A}, \mathcal{T}, \mathcal{R}, \Omega, \mathcal{O} \rangle$. The underlying data values and source code of the target chart are hidden, constituting the unobservable state $s_t \in \mathcal{S}$. Instead, the agent makes decisions based on partial visual observations $o_t = (\mathcal{X}, \mathcal{I}_{t-1}) \in \Omega$, where $\mathcal{I}_{t-1}$ is the chart image compiled from the previously generated code. The agent's action at each step is a composed macro action $u_t = (r_t, a_t) \in \mathcal{A}$. The training objective is to learn a policy $\pi_\theta(u_t \mid h_t)$ (conditioned on the history $h_t$) that maximizes the expected terminal reward: $\theta^* = \arg\max_\theta \mathbb{E}_{\tau \sim \pi_\theta} \left[ R_{\text{final}} \right]$, where the global reward $R_{\text{final}}$ measures the visual consistency between the final rendered chart $\mathcal{I}_K$ and the reference $\mathcal{X}$.

\section{Training Framework}

\begin{figure}[!t]
\centering
\includegraphics[width=1.0\linewidth]{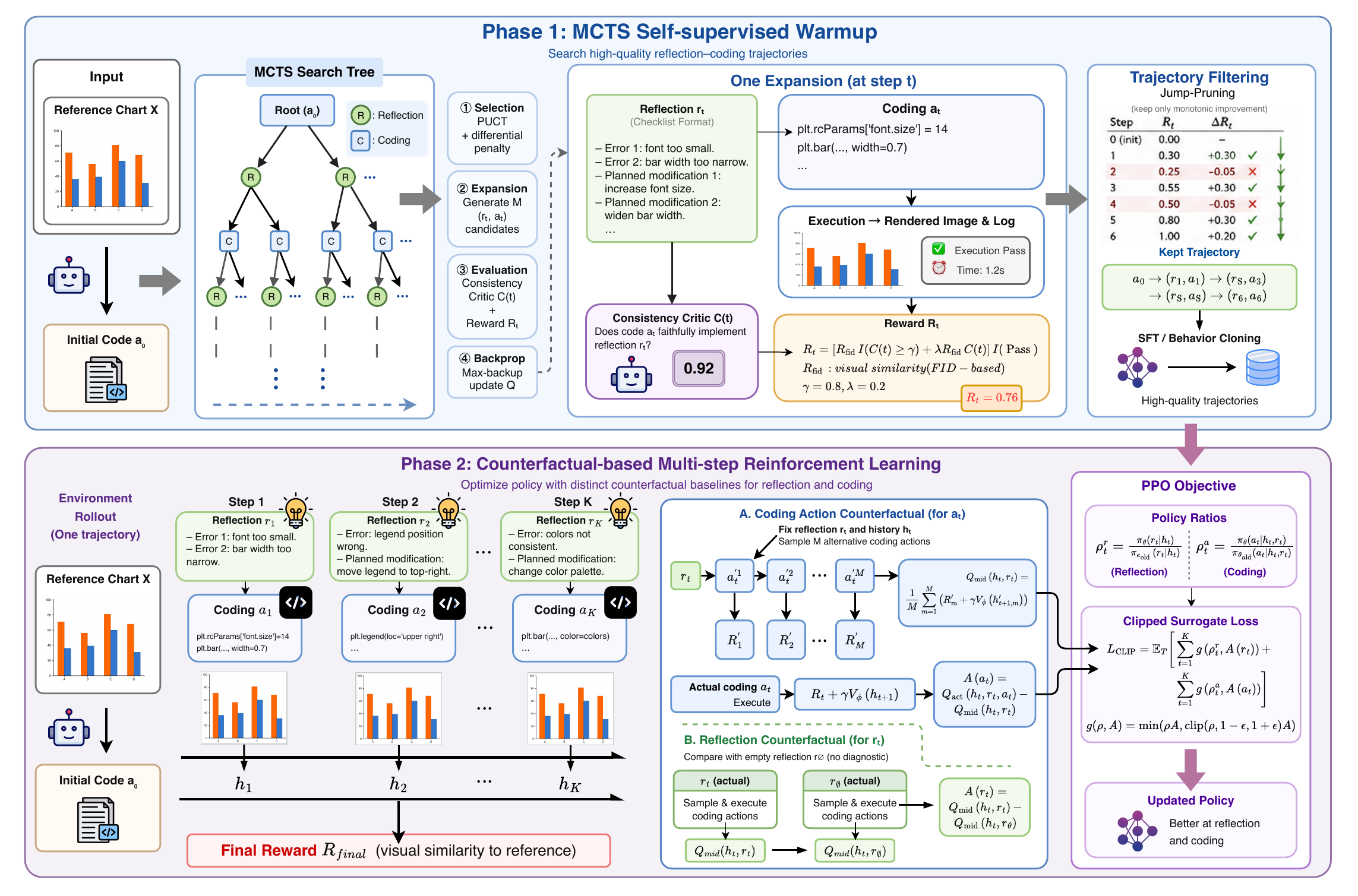}
\caption{Overview of the ViCo two-stage training.}
\label{fig:training}
\end{figure}

This section presents the training framework of ViCo, which consists of a self-supervision-based warming-up stage followed by reinforcement learning for multi-step agentic coding. The overall procedure is demonstrated in Figure~\ref{fig:training} and detailed below.

\subsection{Self-Supervised Warm-Up with MCTS}
To learn a policy that can perform effective reflection during coding, a major challenge is the lack of high-quality reflection–coding trajectories. A common approach is to use a strong LLM as a teacher and generate pseudo-trajectories. However, the reasoning patterns of a strong teacher may differ substantially from those of a weaker student, making direct imitation less effective and potentially disrupting the student's existing reasoning behavior, especially with limited training data. Therefore, we propose a self-supervised warm-up approach that leverages MCTS to synthesize high-quality trajectories through the student's own exploration.
Given a search tree that encodes candidate reflection-coding sequences, a \textbf{consistency-constrained and monotonicity-guaranteed} sampling mechanism is used to prune such branches: 1) the coding step does not faithfully follow the preceding reflection, and 2) iterative refinement fails to improve the rendered quality.

\paragraph{Selection with Differential Penalty:}
Starting from the root node, the algorithm recursively selects child nodes using a variant of PUCT (Predictor Upper Confidence Bound applied to Trees). To avoid wasting computation on branches that have stagnated, we introduce a \textbf{differential penalty term} that penalizes nodes where recent reward improvements are negative.
For a given state \( s_{t-1} \) and candidate action \( u_t = (r_t, a_t) \), the selection score is:
$$
\begin{array}{l}
\text{UCB}^*(s_{t-1}, u_t) = Q(s_{t-1}, u_t) \\
\quad \quad + c \cdot P_{\text{prior}} \cdot \frac{\sqrt{N(s_{t-1})}}{1 + N(s_{t-1}, u_t)} - \mathcal{P}(\Delta R_{\text{history}}),
\end{array}
$$
where \( Q(s_{t-1}, u_t) \) is the maximum reward achieved along this branch, \( N(\cdot) \) counts visits, \( P_{\text{prior}} \) is the prior probability from the policy model, and \( c \) is an exploration constant. The penalty term \( \mathcal{P}(\Delta R_{\text{history}}) \) is defined as:
\[
\mathcal{P}(\Delta R_{\text{history}}) = \beta \cdot \mathbb{I}\{ \text{last } \rho \text{ steps show } \Delta R \leq 0 \},
\]
where \( \beta \) is a scaling coefficient. This mechanism guides search away from dead-end branches.

\paragraph{Expansion with Structured Consistency Constraints:}
At a leaf node, the policy model generates \( M \) candidate child actions \( \{ u_t^{(i)} \}_{i=1}^M \). To allow the quantification of reflection effectiveness, we impose a \textbf{structured output format} through prompt engineering: the reflection \( r_t \) must be expressed as a \textbf{bulleted checklist} of error causes and planned modifications, followed by the code \( a_t \). After generation, each code script is executed in a sandbox environment, and the resulting rendered image, execution logs, and reflection text together form the new state \( s_t^{(i)} \).

\paragraph{Evaluation with Reflection-Consistency Critic and Reward Shaping:}
To address the “say one thing, do another” problem, a lightweight \textbf{Critic} LLM is used to evaluate the consistency between the reflection checklist \( r_t \) and the following codes \( a_t \), outputs a continuous \textbf{consistency score} \( C(t) \in [0,1] \). The reward for the current macro episode is shaped as:
\[
R_t=R_{\mathrm{fid}}(t)(1+\lambda C(t))
\mathbb I(C(t)\ge\gamma)\mathbb I(\mathrm{Pass})
\]
where \( \mathbb{I}(\text{Pass}) \) is the environment test pass rate (1 if execution succeeds, 0 otherwise), \( \gamma \) is a consistency threshold (0.8), and \( \lambda \) is a gain coefficient (0.2). $R_{fid}(t)$ systematically evaluates the fidelity of the rendered code (detailed in Section~\ref{subsubsec:alignments}). If consistency is severely lacking (\( C(t) < \gamma \)), even a successful execution yields zero reward. This discourages superficial successes that ignore the reflection’s guidance. We also implement an \textbf{early stopping mechanism} based on the reward differential \( \Delta R_t = R(t) - R(t-1) \). A patience counter \( \rho \) (initialized to \( \rho_{\max} \)) decrements each time \( \Delta R_t \leq 0 \). When \( \rho \) reaches zero, the branch is labeled a “Dead End” and ceases exploration.

\paragraph{Backpropagation with Max-Backup:}
Given the sparse and binary nature of the final correctness (either the plot matches or it does not), we adopt \textbf{Max-Backup} to propagate the highest reward encountered along a path. For each ancestor node \( i \) on the path from the current leaf to the root:
\[
\begin{aligned}
Q(s_i, u_{i+1}) &\leftarrow \max\big( Q(s_i, u_{i+1}), V \big), \\
N(s_i, u_{i+1}) &\leftarrow N(s_i, u_{i+1}) + 1,
\end{aligned}
\]
where \( V = R_t \) is the leaf's reward. This ensures that high-value branches are quickly prioritized.

\paragraph{Trajectory Extraction:}
Once MCTS completes, we extract the path that yielded the highest final reward \( R_{\text{final}} = 1.0 \) (i.e., a perfect visual match). The raw trajectory inevitably contains intermediate steps with non-positive reward differentials. To distill an efficient correction pattern, we apply \textbf{Jump-Pruning}: remove all macro episodes where \( \Delta R_t \leq 0 \), and concatenate the remaining steps, bridging the environmental feedback from the pruned step to the next retained step.
The resulting trajectories for SFT have a strictly increasing reward sequence:
\[
\tau_{\text{SFT}} = \big( a_0, (r_{t_1}, a_{t_1}), (r_{t_2}, a_{t_2}), \dots, (r_{t^*}, a_{t^*}) \big),
\]
where \( t^* \) is the terminal step. 






\subsection{Counterfactual-based Multi-step RL}
\label{subsec:rl_optimization}
After warmup on self-supervised trajectories, we further employ policy gradient optimization to optimize the expected reward under the POMDP (\S~\ref{sec:2.1}).

\subsubsection{Step-wise Advantage Estimation}
\label{subsubsec:advantage_estimation}

To address the sparse reward problem inherent in multi-step visual replication, we extend the Proximal Policy Optimization (PPO)~\cite{PPO} framework by incorporating a counterfactual-based advantage estimation method inspired by COMA \cite{COMA}. Note that each macro action episode $u_t$ consists of a sequential, non-exchangeable reflection step $r_t$ and coding step $a_t$, we compute advantage for the individual step using differential counterfactual rollouts based on the same history $h_t$.

\paragraph{Advantage of Coding Step $A(a_t)$:} Given the history up to $h_t$ and the chosen reflection $r_t$, the action-value under the current trajectory is:
\[
    Q_{\text{act}}(h_t, r_t, a_t) = R_t + \gamma V_\phi(h_{t+1}),
\]
where $V_\phi$ is the critic network parameterized by $\phi$. To evaluate the quality of $a_t$ under the given reflection, we fix $r_t$ and sample $M$ alternative coding actions $\{ a'_1, \dots, a'_M \}$ from the policy $\pi_\theta(\cdot \mid h_t, r_t)$. Each alternative is executed in the sandbox, producing rewards $R'_m$ and next histories $h'_{t+1,m}$. The expected action-value under the same reflection is approximated through Monte Carlo simulation:
\[
    Q_{\text{mid}}(h_t, r_t) = \frac{1}{M} \sum_{m=1}^M \left( R'_m + \gamma V_\phi(h'_{t+1,m}) \right).
\]
The advantage for the actual coding action is then computed as:
\[
    A(a_t) = Q_{\text{act}}(h_t, r_t, a_t) - Q_{\text{mid}}(h_t, r_t).
\]

\paragraph{Advantage of Reflection Step $A(r_t)$:} To isolate the contribution of the reflection itself, we construct a counterfactual baseline where the reflection is replaced by an \textit{Empty Reflection} $r_\emptyset$ (a prompt instructing the model to directly write code without diagnostic analysis). Under this condition, we sample $M$ coding actions $\{ a^\emptyset_1, \dots, a^\emptyset_M \}$ from $\pi_\theta(\cdot \mid h_t, r_\emptyset)$ and compute the baseline action-value:
\[
    Q_{\text{mid}}(h_t, r_\emptyset) = \frac{1}{M} \sum_{m=1}^M \left( R^\emptyset_m + \gamma V_\phi(h^\emptyset_{t+1,m}) \right).
\]
The reflection advantage is then isolated as:
\[
    A(r_t) = Q_{\text{mid}}(h_t, r_t) - Q_{\text{mid}}(h_t, r_\emptyset).
\]

\subsubsection{Training Objective}
\label{subsubsec:training_objective}

We optimize the policy $\pi_\theta$ using an actor-critic PPO formulation, modified to optimize the sequential action components. The policy parameters $\theta$ are updated by maximizing the clipped surrogate objective using the computed step-wise advantages:
\[
    \mathcal{L}^{\text{CLIP}}(\theta) = \mathbb{E}_\tau \left[ \sum_{t=1}^K \Big( g\big(\rho_t^r, A(r_t)\big) + g\big(\rho_t^a, A(a_t)\big) \Big) \right],
\]
where the standard clipped surrogate loss function:
\[
    g(\rho, A) = \min \big( \rho A, \, \text{clip}(\rho, 1-\epsilon, 1+\epsilon) A \big),
\]
and the action ratios are defined as:
\[
    \rho_t^r = \frac{\pi_\theta(r_t \mid h_t)}{\pi_{\theta_{\text{old}}}(r_t \mid h_t)}, \quad \rho_t^a = \frac{\pi_\theta(a_t \mid h_t, r_t)}{\pi_{\theta_{\text{old}}}(a_t \mid h_t, r_t)},
\]
and $\epsilon$ represents the clip range. 

The critic network $V_\phi$ is updated concurrently by regressing towards the target return using the mean squared error (MSE) loss:
\[
    \mathcal{L}^{\text{VF}}(\phi) = \frac{1}{2} \mathbb{E}_\tau \left[ \sum_{t=1}^K \left( V_\phi(h_t) - \text{Return}_t \right)^2 \right],
\]
where $\text{Return}_t = \sum_{l=t}^K \gamma^{l-t} R_l$ is the discounted cumulative reward from step $t$ onwards. $R_l = 0$ for $l < K$ and $R_K = R_{\text{final}}$ due to the sparse visual alignment reward.

\subsection{Automatic Evaluation based on HHLG}
\label{subsec:evaluation_metric}

Evaluating the fidelity of academic chart replication is fundamentally a structural, semantic, and visual alignment task. To bypass the latency, cost, and lack of interpretability of LLM-as-a-judge approaches, we propose a lightweight automatic framework grounded on a \textbf{Hierarchical Heterogeneous Layout Graph} (HHLG) structure, which significantly facilitates the localization of elements and allows the assessment of their alignment. 

\subsubsection{Hierarchical Data Structure: HHLG}
\label{subsubsec:hhlg}

We use heterogeneous data structures to model the hierarchical layout of complex charts. The global multi-panel layout is modeled as a \textbf{macro-tree}, whereas individual subplots are represented as \textbf{micro-graphs}. Crucially, within each micro-graph, we explicitly separate text elements and visual elements as distinct node categories. 

\paragraph{Macro-Tree (for Global Topology):} We first partition a chart into $N$ distinct subplot regions $\mathcal{S} = \{s_1, s_2, \dots, s_n\}$ using projection profile analysis. We define a Macro-Tree $T$, where the root represents the canvas and its children represent subplots, capturing their global, nested spatial arrangements (e.g., side-by-side, grid, or inset).

\paragraph{Micro-Graph (for Local Topology):} For each subplot $s_k$, we construct a directed Micro-Graph $G_k = (V_k, E_k)$. Nodes within this graph are categorized into two distinct, non-overlapping sets:
\textbf{1) Text Nodes ($V_{\text{text}}$):} Extracted via a high-precision OCR engine - MinerU. These represent title strings, axis labels, tick legends, and in-chart annotations. Each node $u \in V_{\text{text}}$ stores its text content $t_u$ and normalized bounding box $B_u$.
\textbf{2) Visual Nodes ($V_{\text{visual}}$):} Extracted using class-agnostic segmentation anything model~\cite{SAM}. These represent visual data-carrying elements like bars, lines, scatter points, and pie slices. Each node $v \in V_{\text{visual}}$ stores its segmented mask $M_v$ and bounding box $B_v$. For dense datasets (e.g., scatter plots), individual points are grouped into a single cluster node to maintain graph tractability.
\textbf{3) Typed Edges ($E_k$):} Edges are directed and typed to capture both geometric and functional relationships: $E_k \subseteq (V_k \times V_k \times \mathcal{R})$, where the relation type $r \in \mathcal{R}$ includes spatial relations (e.g., $\mathtt{left\text{-}of}$, $\mathtt{above}$) and semantic association relations (e.g., a text node in $V_{\text{text}}$ $\mathtt{labels}$ a visual node in $V_{\text{visual}}$).

\subsubsection{Elements Alignment and Assessment}
\label{subsubsec:alignments}

Once the HHLG is constructed for both the reference chart $x$ and the generated chart $y$, we compute their consistency alongside four dimensions.

\textbf{Semantic Consistency ($S_{\text{sem}}$)} measures the textual alignment between the reference and generated charts. Instead of analyzing the entire image, we isolate the computation to the Text Nodes $V_{\text{text}}$ across matched subplots. For the text nodes of reference $V_{\text{text}, x}$ and generator $V_{\text{text}, y}$, we solve a bipartite matching problem. The pairwise alignment cost $c(u, v)$ between a reference text node $u$ and a generated text node $v$ is defined as:
\[
    c(u, v) = \alpha \cdot d_{\text{edit}}(t_u, t_v) + (1-\alpha) \cdot \left(1 - \text{IoU}(B_u, B_v)\right)
\]
where $d_{\text{edit}}$ is the normalized Levenshtein distance of the strings, and $\text{IoU}$ measures the spatial bounding box intersection-over-union. The final semantic score $S_{\text{sem}}$ is computed via the Hungarian algorithm to minimize global alignment cost, normalized by the node count.

\textbf{Layout Consistency ($S_{\text{lay}}$)} evaluates the spatial and structural preservation of the chart layout at both the macro and micro levels:
\[
    d_{\text{layout}}(x, y) = d_{\text{Tree}}(T_x, T_y) + \sum_{k} \text{GED}(G_{x,k}, G_{y,\pi(k)})
\]
where $d_{\text{Tree}}$ is the tree edit distance matching the macro subplot layouts, and $\pi(k)$ represents the optimal subplot mapping. The micro-layout discrepancy $\text{GED}$ computes the Graph Edit Distance between matched subplot micro-graphs. Because nodes are explicitly categorized into $V_{\text{text}}$ and $V_{\text{visual}}$, the node substitution cost in GED is constrained: a text node cannot match a visual node. This categorization significantly prunes the search space, allowing us to use a fast, linear-time bipartite approximation of GED without sacrificing evaluation accuracy. The final score is normalized as:
\[
    S_{\text{lay}} = \exp(-\beta \cdot d_{\text{layout}})
\]

\textbf{Style Consistency ($S_{\text{style}}$)}.
To evaluate styles like font faces, gridline textures, and rendering strokes without interference from global spatial layout shifts, we extract visual style features locally. For each matched subplot node in the Macro-Tree, we extract the image patch bounding the subplot. We pass these patches through a convolutional backbone (i.e., VGG-19~\cite{vgg}) to extract activation feature maps $F^l$ at intermediate layers. We then calculate style via the Gram Matrix $G^l$, which discards spatial configurations and focuses strictly on texture and stylistic patterns:
\[
    G^l_{i,j} = \frac{1}{H_l W_l} \sum_{m=1}^{H_l W_l} F^l_{i,m} F^l_{j,m}
\]
The style score is determined by the distance between the Gram matrices of matched subplot patches:
\[
    S_{\text{style}} = \exp \left( -\gamma \sum_{l \in L} w_l \|G^l_x - G^l_y\|_F^2 \right)
\]
where $\|\cdot\|_F$ denotes the Frobenius norm.

\textbf{Color Consistency ($S_{\text{color}}$)}.
To prevent incorrect color-to-entity mappings (e.g., swapping colors between a bar and a line), color consistency is calculated directly on the Visual Nodes $V_{\text{visual}}$. For each visual node $v \in V_{\text{visual}}$ in the micro-graphs, we extract the dominant color vector $\mathbf{c}_v$ in the perceptually uniform LAB color space from its segmented mask $M_v$ (excluding backgrounds). During micro-graph alignment, we compute the color distance between aligned visual node pairs. The color consistency score is defined as:
\[
    S_{\text{color}} = \frac{1}{|M|} \sum_{(u, v) \in M} \left( 1 - \frac{\|\mathbf{c}_u - \mathbf{c}_v\|_2}{\Delta E_{\text{max}}} \right)
\]
where $M$ is the set of successfully matched visual node pairs $(u, v)$ from the layout alignment step, and $\Delta E_{\text{max}}$ represents the maximum possible perceptual color difference. This guarantees that color evaluation is explicitly bound to specific visual elements.

\textbf{Composite Metric}. The final multi-dimensional evaluation score $R$ is formulated as a weighted linear combination of the four graph-grounded metrics: $R_{fid} = w_{1} S_{\text{sem}} + w_{2} S_{\text{lay}} + w_{3} S_{\text{style}} + w_{4} S_{\text{color}}$, where $\sum_{i=1}^{4} w_i = 1$. By grounding all four dimensions in the hierarchical graph, this metric provides stable, highly interpretable, and computationally efficient signals for our two-stage training.

\section{Experiments}
\label{sec:experiments}

\subsection{Benchmarks}
\label{subsec:setup}

Three challenging benchmarks are used in our experiments: \textit{RealChart2Code}~\cite{RealChart2Code}, \textit{ChartMimic}~\cite{shi2024chartmimic}, and \textit{Plot2Code}~\cite{wu2024plot2code}.  
Appendix~\ref{sec:appendix_bench} presents their details and specialized metrics.




\begin{table*}[!htbp]
\centering
\caption{
Evaluation results on three benchmarks. 
*: Only the chart replication task is evaluated. 
The best score within each category is \textbf{bolded} and the second-best is \underline{underlined}.
\dag: Customized replication is achieved through a post-hoc data refinement.
($\uparrow$): the absolute performance gain of ViCo against the Qwen3-VL base models (4B/8B).
}
\label{tab:merged_results}
\resizebox{\textwidth}{!}{
\begin{tabular}{lc cc cccc ccc}
\toprule
\multirow{2.5}{*}{\textbf{Model}} & \multirow{2.5}{*}{\textbf{Size}} & \multicolumn{2}{c}{\textbf{RealChart2Code}$^{*}$} & \multicolumn{4}{c}{\textbf{ChartMimic}} & \multicolumn{3}{c}{\textbf{Plot2Code}} \\
\cmidrule(lr){3-4} \cmidrule(lr){5-8} \cmidrule(lr){9-11}
& & Pass (\%) & Score & Direct P.(\%) & Direct S. & Cust. P$^{\dag}$.(\%) & Cust. S$^{\dag}$. & Pass (\%) & Text & Rating \\
\midrule
\multicolumn{11}{c}{\textit{Closed-source LLMs}} \\
\midrule
Claude-4.5-Sonnet~\cite{anthropic2025claudesonnet45} & - & \underline{85.8} & 7.1 & \textbf{100.0} & 90.1 & \underline{99.5} & 91.5 & 93.9 & 73.3 & 8.8 \\
Claude-4.5-Opus~\cite{anthropic2025claudesonnet45}    & - & \textbf{87.7} & \underline{7.8} & \underline{98.5} & \underline{92.3} & \textbf{100.0} & \underline{95.1} & \textbf{99.6} & 75.4 & \underline{9.0} \\
Gemini-2.5-Flash~\cite{deepmind2026gemini3}          & - & 63.2 & 5.2 & 88.5 & 69.8 & 89.1 & 74.4 & 87.9 & 72.8 & 8.6 \\
Gemini-3-Pro-Preview~\cite{deepmind2026gemini3}      & - & 82.1 & \textbf{9.0} & 97.3 & \textbf{96.0} & \textbf{100.0} & \textbf{96.8} & 90.9 & \underline{79.5} & \textbf{9.6} \\
GPT-5.1~\cite{singh2026openaigpt5card}               & - & 71.2 & 5.7 & 97.8 & 89.8 & 98.5 & 91.2 & \underline{98.5} & \textbf{82.6} & 8.8 \\
\midrule
\multicolumn{11}{c}{\textit{Open-source LLMs}} \\
\midrule
DeepSeek-VL~\cite{DeepSeekVL}   & 7B   & 9.7 & 0.4 & 41.3 & 19.7 & 59.3 & 37.6 & 64.4 & 32.6 & 2.3 \\
Intern-VL-3.5~\cite{InternVL3}  & 241B & \textbf{54.3} & \textbf{3.5} & 90.5 & \underline{76.9} & 90.7 & \underline{78.4} & \textbf{90.2} & \underline{69.0} & 7.6 \\
Intern-VL-3.5~\cite{InternVL3}  & 30B  & 20.3 & 0.8 & 85.6 & 68.1 & 87.2 & 68.5 & 88.6 & 66.3 & 6.7 \\
Qwen3-VL~\cite{Qwen3VL}         & 235B & \underline{49.2} & 3.3 & \textbf{91.5} & \textbf{80.4} & \textbf{92.7} & \textbf{81.8} & 89.8 & \textbf{69.8} & \textbf{8.9} \\
Qwen3-VL~\cite{Qwen3VL}         & 30B  & 17.3 & 0.8 & 89.5 & 68.6 & 89.1 & 69.9 & 88.9 & 65.0 & \underline{7.9} \\
GLM-4.5V~\cite{GLM45V}          & 106B & 38.4 & 2.8 & 87.4 & 72.8 & 86.7 & 71.2 & 84.1 & 61.7 & 5.8 \\
GLM-4.1V~\cite{GLM45V}          & 9B   & 12.8 & 0.7 & 86.6 & 70.2 & 87.1 & 70.1 & 74.1 & 59.1 & 4.9 \\
MiMo-VL-RL~\cite{MiMoVL}        & 7B   & 11.0 & 0.4 & 83.2 & 58.5 & 87.7 & 64.7 & 73.5 & 55.5 & 4.3 \\
ChartCoder~\cite{ChartCoder}    & -    & 48.0 & \textbf{3.5} & \underline{90.9} & 75.3 & \underline{92.1} & 76.7 & 87.9 & 54.5 & 4.7 \\
\midrule
\midrule
\rowcolor{gray!10} ViCo (Qwen3-VL) & 4B & 98.4\small($\uparrow$18.5) & 3.2\small($\uparrow$1.4) & 98.1\small($\uparrow$14.2) & 78.8\small($\uparrow$12.5) & 100.0\small($\uparrow$16.8) & 79.2\small($\uparrow$15.0) & 97.2\small($\uparrow$18.6) & 69.5\small($\uparrow$12.3) & 8.1\small($\uparrow$2.5) \\
\rowcolor{gray!10} ViCo (Qwen3-VL) & 8B & 98.8\small($\uparrow$12.2) & 3.4\small($\uparrow$0.9) & 99.2\small($\uparrow$9.5) & 81.5\small($\uparrow$8.2) & 100.0\small($\uparrow$7.5) & 80.9\small($\uparrow$6.8) & 99.1\small($\uparrow$10.3) & 78.4\small($\uparrow$8.9) & 8.4\small($\uparrow$1.6) \\
\bottomrule
\end{tabular}
}
\end{table*}

\subsection{Main Results}
\label{subsec:main_results}
Table~\ref{tab:merged_results} compares ViCo against prevalent closed- and open-source models (Scores of the compared models are reported in~\cite{RealChart2Code}).
Results demonstrate that ViCo achieves outstanding code compilation reliability, consistently outperforming both open- and closed-source models in \textbf{Pass Rate (Pass \%)}. 

On RealChart2Code, ViCo (8B) secures a \textbf{98.8\%} pass rate, surpassing Claude-4.5-Opus (87.7\%) and Intern-VL-3.5-241B (54.3\%). It also reaches \textbf{100.0\%} on ChartMimic (Customized) and \textbf{99.1\%} on Plot2Code. This represents a substantial improvement over the Qwen3-VL base models, showing absolute gains ($\uparrow$) of up to 18.6\% in pass rates and significant boosts across visual/textual scores (e.g., a +2.5 rating gain for ViCo-4B on Plot2Code).
Furthermore, despite its compact scale, ViCo's alignment scores closely approach those of state-of-the-art closed-source models. 
On ChartMimic, ViCo (8B) obtains a Direct Score of \textbf{81.5} (surpassing Gemini-2.5-Flash's 69.8). 
On Plot2Code, it achieves a Text score of \textbf{78.4} and an 8.4 Rating, which are highly competitive with Gemini-2.5-Flash (72.8 and 8.6, respectively) and Claude-4.5-Sonnet (73.3 and 8.8). 
While evaluation scores on RealChart2Code are generally low across all evaluated models due to the challenges of complex multi-subplot structures, ViCo (8B) remains highly competitive with a \textbf{3.4} point (against ChartCoder's 3.5). Our ablation study ($\S$~\ref{sec:ablation}) attributes the degraded performance on multi-subplots primarily to the bottleneck of the visual rewarding mechanism rather than generation failure.

\subsection{Ablation Studies}
\label{sec:ablation}

We conduct ablation studies using ViCo (8B) variants to evaluate the individual contributions of our core components. Table~\ref{tab:ablation_results} summarizes the performance across different configurations.

\paragraph{HHLG vs. Open LLM-As-A-Judge.}
We select 200 samples from RealChart2Code and ChartMimic, categorizing them into simple layouts (no subplots or $1\times2$ subplots) and complex structures ($2\times2$ subplots or above) for a correlation analysis against human judgment.  As shown in Table~\ref{tab:human_corr}, our HHLG-based metric exhibits a stronger correlation with human judgment across both simple ($\rho = 0.724$) and complex ($\rho = 0.458$) settings. 
This higher alignment with human explains the ablation results in Table~\ref{tab:ablation_results} (Group 1): replacing HHLG with Qwen3-VL-8B as the reward model degrades the visual score from 81.5 to 72.3 on ChartMimic and from 3.4 to 2.8 on RealChart2Code. 
Additionally, the correlation scores of all three reward mechanisms with human preferences decrease sharply on complex layouts (with all Spearman $\rho < 0.5$). This uniform decline reveals that providing precise, automatic visual reward signals for dense, multi-subplot configurations remains a persistent challenge for current models. 
This reward-utility bottleneck on complex layouts is the primary reason why ViCo's visual score improvement on RealChart2Code.

\begin{figure*}[!ht]
\centering
\begin{subfigure}[b]{0.33\linewidth}
\centering
\includegraphics[width=\linewidth]{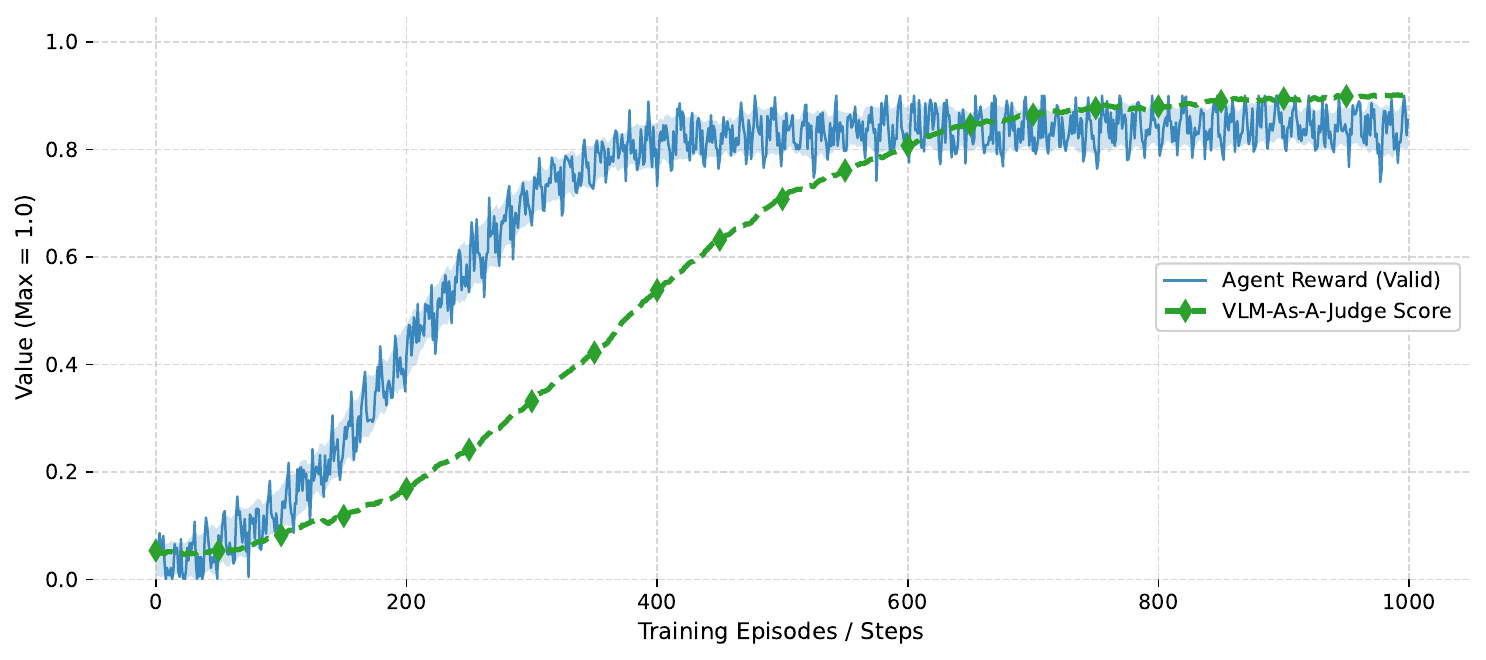}
\caption{Our original PPO}
\label{subfig:ppo_ours}
\end{subfigure}%
\begin{subfigure}[b]{0.33\linewidth}
\centering
\includegraphics[width=\linewidth]{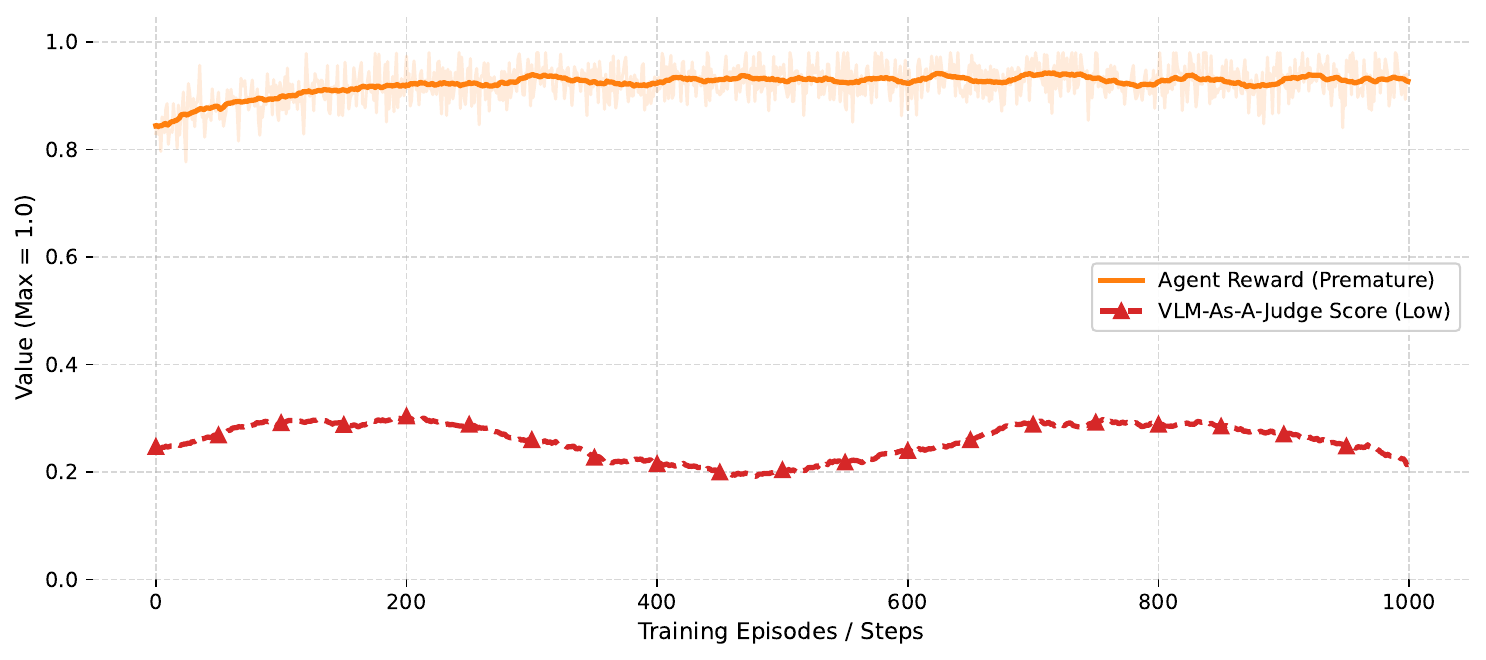}
\caption{Our PPO w/o HHLG-based reward}
\label{subfig:ppo_premature}
\end{subfigure}%
\begin{subfigure}[b]{0.33\linewidth}
\centering
\includegraphics[width=\linewidth]{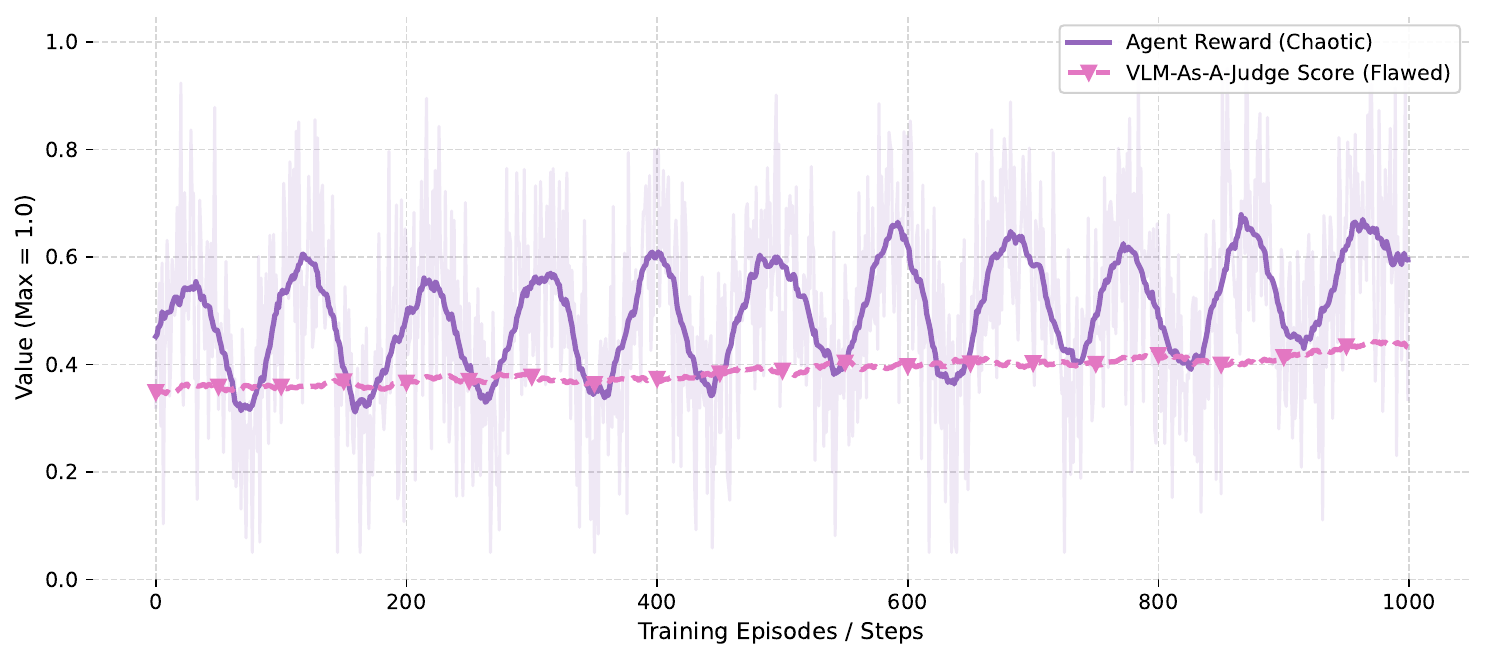}
\caption{Our PPO w/o refined credit assignment}
\label{subfig:ppo_no_credit}
\end{subfigure}
\caption{
Training dynamics of multi-step PPO with different rewarding and credit assignment mechanisms.
}
\label{fig:ppo_training_comparison}
\end{figure*}

\begin{table}[!t]
\centering
\caption{Ablation studies of ViCo (8B) configurations evaƒluated on RealChart2Code and ChartMimic (Direct Mimic). We report the Pass Rate (\%) and visual Score.}
\label{tab:ablation_results}
\resizebox{\columnwidth}{!}{
\begin{tabular}{l|cc|cc}
\toprule
\multirow{2}{*}{\textbf{Configuration}} & \multicolumn{2}{c|}{\textbf{RealChart2Code}} & \multicolumn{2}{c}{\textbf{ChartMimic}} \\
 & Pass (\%) & Score & Pass (\%) & Score \\
\midrule
\rowcolor{gray!10} ViCo (8B) (\textbf{Full Framework}) & \textbf{98.8} & \textbf{3.4} & \textbf{99.2} & \textbf{81.5} \\
w/o RL (Warmup-only) & 96.5 & 3.2 & 94.2 & 77.1  \\
w/o SFT (Cold-start RL) & 50.5 & 2.7 & 69.9	& 69.7 \\
\midrule
\rowcolor{gray!10} \multicolumn{5}{l}{\textit{Rewarding Strategy (Sec.~\ref{sec:ablation}, Group 1)}} \\
HHLG $\to$ LLM-As-A-Judge & 94.1 & 2.8 & 95.0 & 72.3 \\
\midrule
\rowcolor{gray!10} \multicolumn{5}{l}{\textit{MCTS Filtering \& Constraints (Sec.~\ref{sec:ablation}, Group 2)}} \\
\quad w/o Consistency Checking & 96.3 & 3.1 & 98.5 & 76.4 \\
\quad w/o Format Constraints & 96.1 & 3.2 & 99.4 & 78.1 \\
\quad w/o Jump-Pruning & 96.5 & 3.2 & 89.1 & 77.8 \\
\midrule
\rowcolor{gray!10} \multicolumn{5}{l}{\textit{RL Credit Assignment (Sec.~\ref{sec:ablation}, Group 3)}} \\
\quad w/o Counterfactual (Multi-turn PPO) & 47.1 & 3.0 & 87.2 & 75.1 \\
\quad w/o Reflection Credit & 49.0 & 3.1 & 88.9 & 77.0 \\
\bottomrule
\end{tabular}
}
\end{table}

\begin{table}[!t]
\centering
\small
\caption{Correlation between different evaluation metrics and human judgment, categorized by chart complexity (\textit{Sim.}: Simple, \textit{Com.}: Complex). 
}
\label{tab:human_corr}
\resizebox{\linewidth}{!}{
\begin{tabular}{lcccccc}
\toprule
\multirow{2.5}{*}{\textbf{Evaluation Metric}} & \multicolumn{2}{c}{\textbf{Spearman ($\rho$)}} & \multicolumn{2}{c}{\textbf{Pearson ($r$)}} & \multicolumn{2}{c}{\textbf{Kendall ($\tau$)}} \\
\cmidrule(lr){2-3} \cmidrule(lr){4-5} \cmidrule(lr){6-7}
& \textit{Sim.} & \textit{Com.} & \textit{Sim.} & \textit{Com.} & \textit{Sim.} & \textit{Com.} \\
\midrule
\rowcolor{gray!10} \multicolumn{7}{l}{\textit{Baselines (LLM-as-a-Judge)}} \\[4pt]
Qwen3-VL (8B) & 0.425 & 0.184 & 0.451 & 0.203 & 0.318 & 0.125 \\
GPT-4o        & \underline{0.712} & \underline{0.321} & \textbf{0.738} & \underline{0.355} & \textbf{0.546} & \underline{0.241} \\
\midrule
\textbf{HHLG-Based (Ours)} & \textbf{0.724} & \textbf{0.458} & \underline{0.701} & \textbf{0.412} & \underline{0.503} & \textbf{0.317} \\
\bottomrule
\end{tabular}
}
\end{table}

\paragraph{MCTS Consistency Constraints.}
We isolate the components of our MCTS-based trajectory sampling framework (Table~\ref{tab:ablation_results}, Group 2). Excluding the \textit{consistency critic check} results in a drop in visual score (from 81.5 to 76.4 on ChartMimic and 3.4 to 3.1 on RealChart2Code), indicating the model occasionally ignores its own planned corrections. 
Removing \textit{format constraints} degrades performance across both benchmarks (e.g., dropping to 78.1 on ChartMimic). Lastly, removing jump-pruning degrades trajectory quality and execution success, causing the code pass rate on ChartMimic to drop to 89.1\%. 
This highlights the necessity of keeping search trajectories focused on strictly positive refinement actions.

\paragraph{Counterfactual-based Credit Assignment.}
As demonstrated in Table~\ref{tab:ablation_results} (Group 3), vanilla multi-turn PPO (without counterfactual baselines) suffers from a severe performance decline: the RealChart2Code pass rate plunging from 98.8\% to 47.1\%, while omitting reflection credit drops the pass rate to 49.0\%. 
These steep declines are visually explained by the training curves in Figure~\ref{fig:ppo_training_comparison}. 
Under our full framework (Figure~\ref{subfig:ppo_ours}), the agent reward and the actual visual evaluation score scale synchronously, steadily converging to a high level. This indicates a robust mapping between visual critiques ($r_t$) and coding edits ($a_t$). 
In contrast, when using standard LLM-as-a-judge rewards without HHLG (Figure~\ref{subfig:ppo_premature}), the policy quickly exploits weaknesses in the reward model (reward hacking), leading to mismatched training dynamics where agent rewards spike prematurely to near-maximum while the true evaluation scores remain flat and low. 
Furthermore, removing our credit assignment mechanism (Figure~\ref{subfig:ppo_no_credit}) causes severe credit blurring under sparse terminal rewards. This leads to highly chaotic, periodic fluctuations in agent rewards throughout training, with no substantive gains in visual evaluations. 

\begin{table}[!t]
\centering
\small
\caption{Performance breakdown across layout complexity tiers on RealChart2Code.
}
\label{tab:subplot_complexity}
\resizebox{\columnwidth}{!}{
\begin{tabular}{l cc cc cc}
\toprule
\multirow{2.5}{*}{\textbf{Model}} & \multicolumn{2}{c}{\textbf{Simple ($1\times 1$)}} & \multicolumn{2}{c}{\textbf{Medium ($2\times 2$--$2\times 3$)}} & \multicolumn{2}{c}{\textbf{Complex ($>6$ subplots)}} \\
\cmidrule(lr){2-3} \cmidrule(lr){4-5} \cmidrule(lr){6-7}
& Pass (\%) & Score & Pass (\%) & Score & Pass (\%) & Score \\
\midrule
Qwen3-VL-8B      & 90.8 & 5.8 & 78.8 & 2.8 & 20.3 & 0.5 \\
ChartCoder       & 89.5 & 5.6 & 72.3 & 3.5 & 28.0 & 0.7 \\
\midrule
ViCo (8B)           & \textbf{100.0} & \textbf{8.7} & 99.0 & 6.7 & 99.0 & 1.2 \\
\rowcolor{gray!10} \quad \textit{+ Divide-and-Conquer} & -- & -- & \textbf{100.0} & \textbf{8.2} & \textbf{100.0} & \textbf{7.6} \\
\bottomrule
\end{tabular}
}
\end{table}

\section{Discussions}
\subsection{Performance on Complex Layouts}
\label{subsec:multi_subplot_analysis}

To investigate the effect of structural density on replication fidelity, we evaluated 200 sampled RealChart2Code charts categorized across three complexity tiers: \textit{Simple} ($1\times 1$), \textit{Medium} ($2\times 2$--$2\times 3$), and \textit{Complex} ($>6$ subplots, e.g., $3\times 3$ grids). As shown in Table~\ref{tab:subplot_complexity}, end-to-end ViCo-8B sustains near-perfect code execution ($\ge 99.0\%$) and delivers substantial visual score gains on Simple ($8.7$ vs. $5.8$) and Medium ($6.7$ vs. $2.8$) tiers over the base model. However, direct end-to-end generation on dense Complex grids degrades to a score of $1.2$ across all models due to visual grounding and resolution bottlenecks. 
\textbf{Remedy:} To overcome this limitation, we developed a modular \textbf{Divide-and-Conquer pipeline} that decomposes multi-panel figures into individual subplots via bounding-box projection analysis, replicates each panel independently with ViCo, and programmatically stitches them back into the detected canvas grid. This strategy drastically elevates the Complex tier visual score from $1.2$ to $\mathbf{7.6}$ (and Medium to $\mathbf{8.2}$) with a $100.0\%$ pass rate, demonstrating a practical and robust pathway for deploying ViCo on high-density academic charts.




\subsection{Cross-Backbone Generalization}
\label{subsec:cross_backbone}

\begin{figure}
\centering
\includegraphics[width=0.95\linewidth]{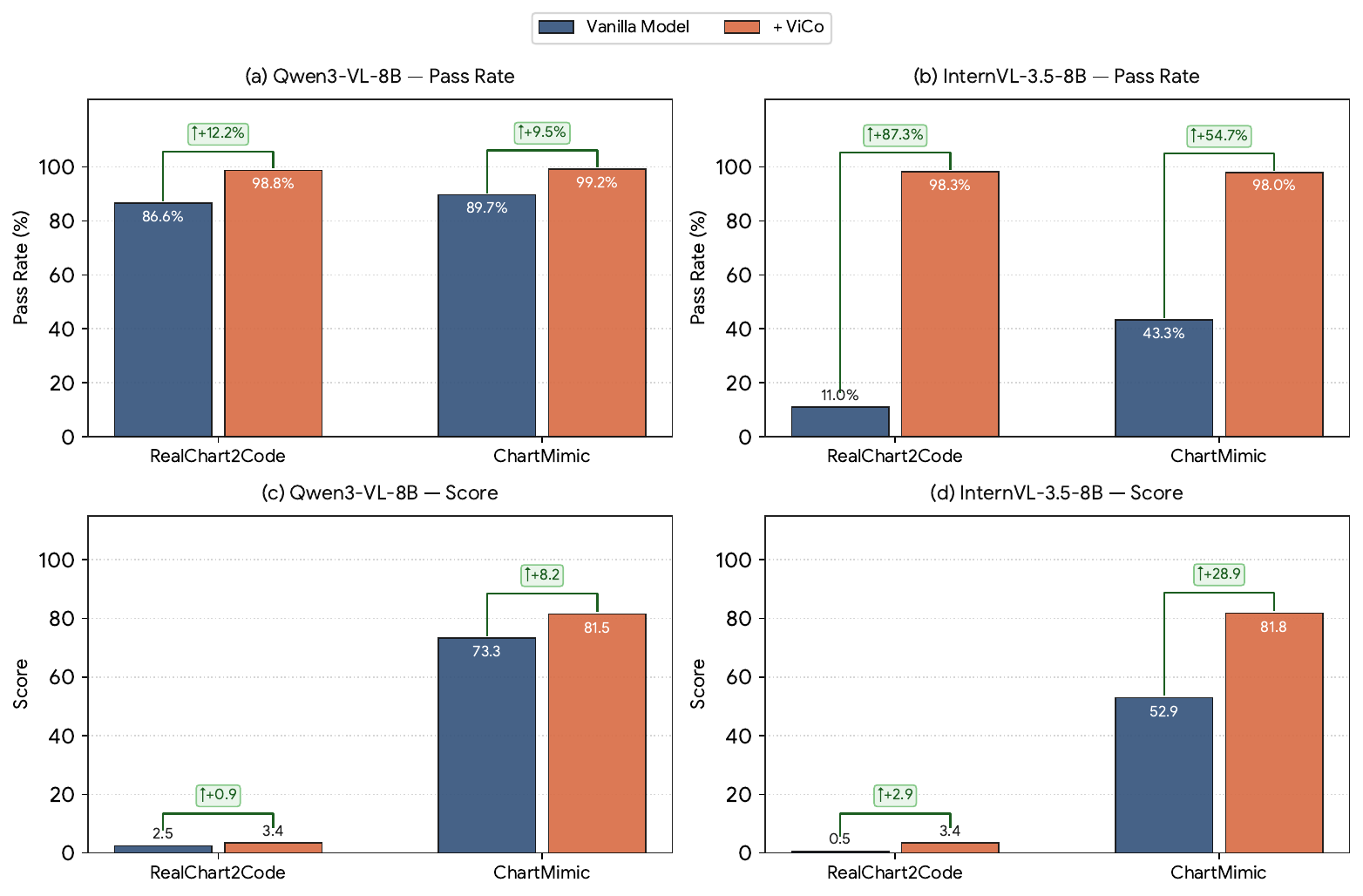}
\caption{Cross-backbone generalization of ViCo evaluated on RealChart2Code and ChartMimic.}
\label{fig:cross_backbone}
\end{figure}

We instantiate and evaluate ViCo on another representative open-source multimodal backbone, \textbf{Intern-VL-3.5-8B}, alongside our default backbone \textbf{Qwen3-VL-8B}. As shown in Figure~\ref{fig:cross_backbone}, the native InternVL-3.5-8B exhibits severe execution and visual alignment deficiencies on code generation. Integrating ViCo yields drastic, across-the-board performance leaps: code pass rates surge to $\mathbf{98.3\%}$ ($+87.3\%$) on RealChart2Code and $\mathbf{98.0\%}$ ($+54.7\%$) on ChartMimic, while visual fidelity scores jump to $\mathbf{3.4}$ and $\mathbf{81.8}$ ($+28.9\%$), respectively, matching the competitive performance of ViCo on Qwen3-VL-8B. These consistent, decisive gains across distinct vision-language architectures confirm that ViCo's structured warm-up and counterfactual credit allocation mechanisms are universally transferable rather than reliant on specific model priors.


\subsection{Effectiveness of Self-Reflection}
\begin{figure}[!t]
\centering
\begin{subfigure}[b]{0.95\linewidth}
\centering
\includegraphics[width=\linewidth]{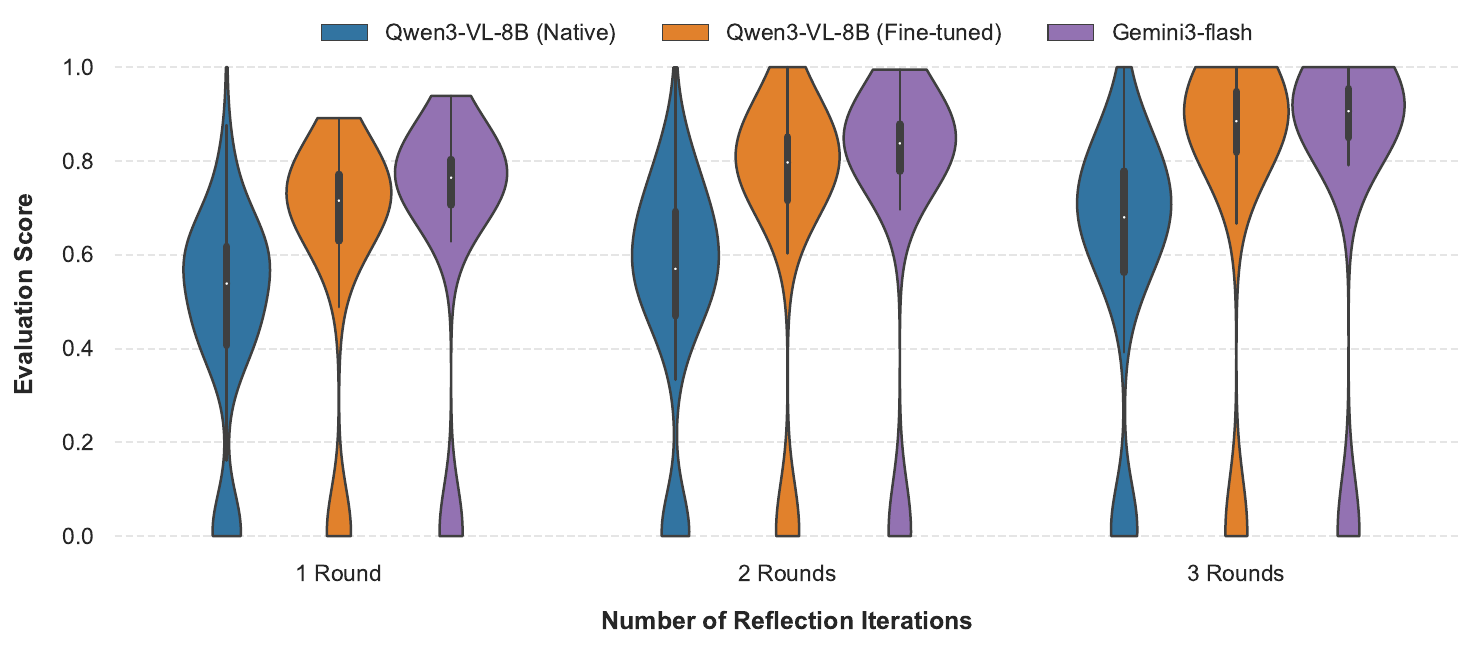} 
\caption{Comparison of GPT-4o evaluation scores.}
\label{fig:reflection_reward}
\end{subfigure}
\begin{subfigure}[b]{0.95\linewidth}
\centering
\includegraphics[width=\linewidth]{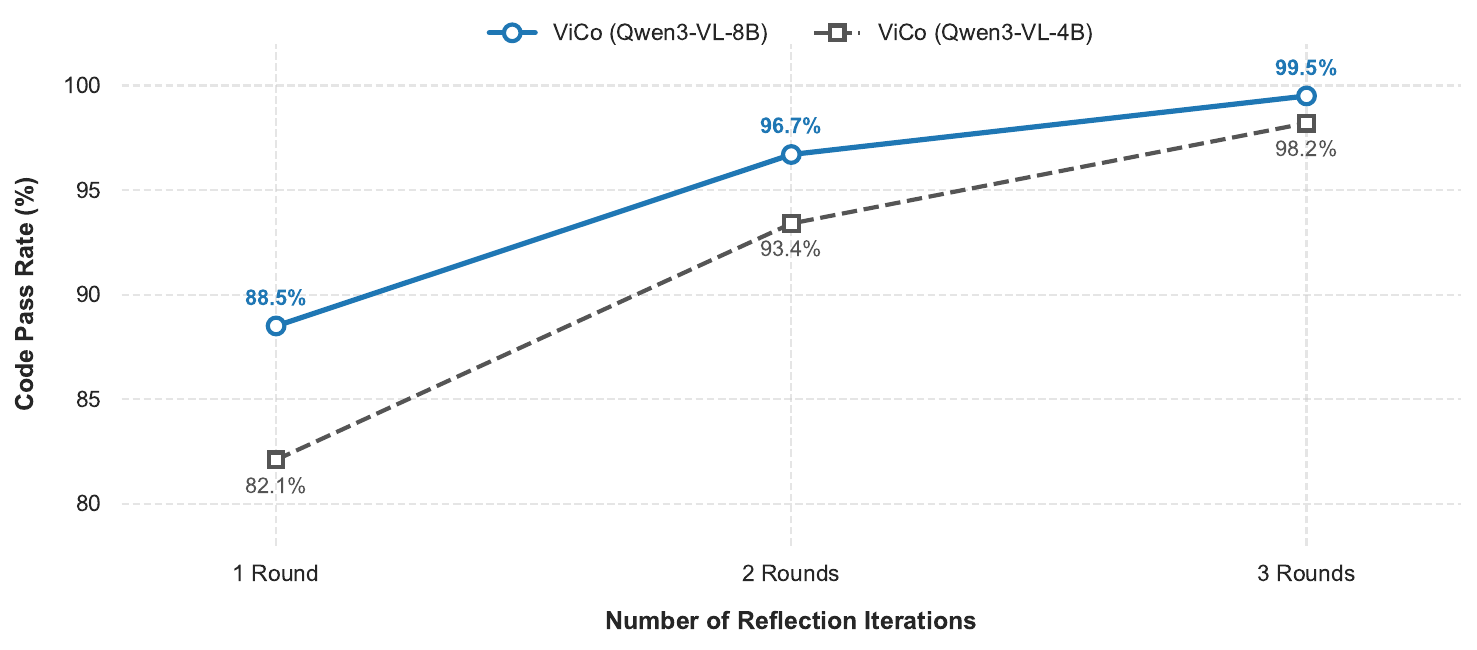} 
\caption{Comparison of code pass rate.}
\label{fig:reflection_pass}
\end{subfigure}
\caption{Evaluation against reflection iterations.}
\label{fig:reflection}
\end{figure}

We analyze the performance of ViCo up to three refinement rounds on 200 ChartMimic samples. 
In Figure~\ref{fig:reflection_reward}, the score distribution of Qwen3-VL-8B remains stagnant and widely dispersed across rounds, ViCo steadily shifts its evaluation scores toward the top range (0.8--1.0) by Round 3, closely matching the trajectory of the proprietary Gemini3-flash. 
Figure~\ref{fig:reflection_pass} show a monotonic scaling of code pass rates over iterations. This proves that iterative self-reflection, guided by our consistency constraints, systematically resolves execution errors and progressively refines visual alignment.

\subsection{Budget of Counterfactual Sampling}
\label{subsec:baseline_scaling}
\begin{figure}[!t]
\centering
\includegraphics[width=0.95\linewidth]{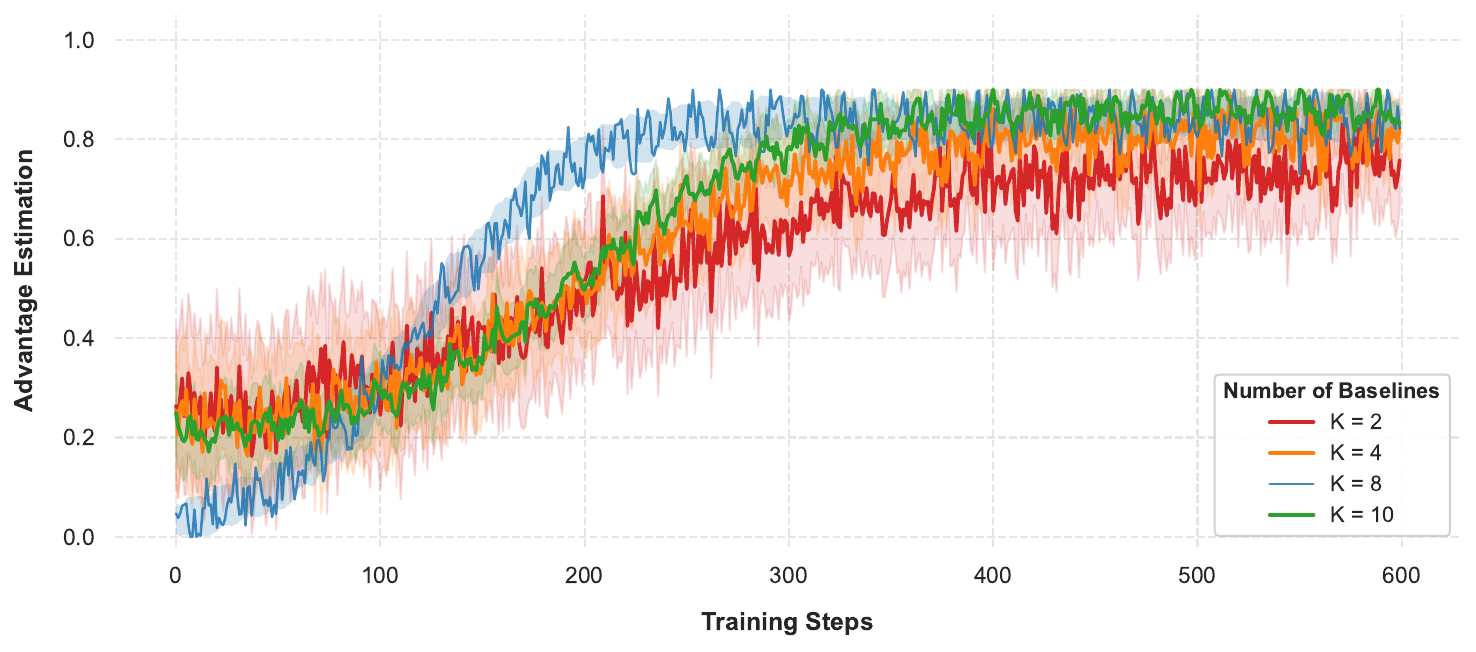}
\caption{Comparison of advantage estimation in PPO with varying number of counterfactual baselines.}
\label{fig:baselines}
\end{figure}

We analyze the variance of the counterfactual advantage estimators $A(r_t)$ and $A(a_t)$ across varying sample budgets $M \in \{2, 4, 8, 10\}$. As shown in Figure~\ref{fig:baselines}, scaling $M$ from $2$ to $8$ drastically shrinks gradient variance and prevents destructive policy updates caused by noisy Monte Carlo approximations. Beyond $M=8$, the marginal variance reduction becomes negligible while the wall-clock sandbox execution time scales linearly. This confirms that $M=8$ strikes an optimal balance, providing sufficiently dense and low-variance credit assignment signals without incurring unnecessary computational overhead during RL training.

\section{Conclusion}
This paper presents ViCo, a visual-oriented coding framework, which replicates high-quality academic charts through iterative self-reflection. By cascading self-supervised warm-up and counterfactual-based reinforcement learning, ViCo addresses the ineffectiveness of self-critique and the sparse reward problem in multi-turn refinement. 
We also propose HHLG, which enables multifaceted reward encompassing chart quality - style, layout, and semantic alignment. 
With 8B parameters, our trained ViCo models achieves performance close to proprietary LLMs with reflection capabilities across widely used benchmarks.

\section*{Limitations}
\label{sec:limitations}

Despite the performance improvements demonstrated by ViCo, several limitations remain to be addressed in future work.
\begin{itemize}[leftmargin=*]
\item First, while ViCo achieves strong results on standard replication benchmarks, \textbf{complex multi-subplot structures remain a significant challenge}. As observed in our experiments on RealChart2Code, multi-panel layouts (such as dense $2\times2$ and $3\times3$ multiple subplots) involve intricate geometric alignments and highly concentrated visual features. These complex arrangements pose severe difficulties for both precise visual-oriented code generation and accurate automatic evaluation. Resolving these challenges likely requires more sophisticated visual-spatial modeling and layout-aware reasoning paradigms.
\item Second, \textbf{the computational overhead during RL training is relatively high}. Because our counterfactual-based credit assignment mechanism requires curating separate baselines for both the reflection and coding actions at each refinement step, the policy optimization process demands multiple parallel rollouts and counterfactual evaluations. While this design is essential for providing dense, step-level training signals and stabilizing RL without an expensive process reward model, it introduces additional computational burden and training time compared to standard multi-step RL algorithms.
\item Third, \textbf{the scope of our study is limited to replicating academic chart}. Although our core methodology - encompassing reflective self-training and hierarchical visual layout alignment - is generalizable to a broader spectrum of visual-to-code generation tasks, we have not yet evaluated its efficacy on other complex domains. These include web page replication, user interface (UI) reconstruction, or general visual design translation. Validating and adapting ViCo to these wider, general-purpose coding benchmarks remains an important direction for our future research.
\end{itemize}

\newpage
\bibliography{custom}

\appendix
\section{Related Work}

\subsection{Code Generation from Visual Inputs}
The task of translating visual representations (e.g., charts, UI screenshots) into executable code has garnered significant attention. ChartCoder~\cite{ChartCoder} pioneered the chart-to-code paradigm by leveraging a Code LLM backbone and the Snippet-of-Thought (SoT) method. Its training relies on static supervised fine-tuning without visual outcome feedback. VinciCoder~\cite{VinciCoder} proposed a unified multimodal code generation model with a coarse-to-fine visual reinforcement learning (ViRL) strategy. VisCodex~\cite{VisCodex} uses task vector merging to combine vision and coding models. More closely related to us are methods that incorporate visual feedback for code refinement. VisRefiner~\cite{VisRefiner} introduces a training framework where models learn from visual differences between rendered predictions and target designs, followed by a reinforcement learning stage for self-refinement. However, VisRefiner operates on discrete refinement steps without an explicit structured reflection mechanism that enforces consistency between critique and subsequent code changes. In contrast, our proposed ViCo framework employs a dedicated reflection step with consistency constraints, ensuring that every coding action strictly follows the preceding reflection, thereby addressing the ``say one thing, do another'' problem.

\subsection{Reinforcement Learning with Visual Feedback for Code Generation}
Reinforcement Learning with Verifiable Rewards (RLVR)~\cite{RLVRa,RLVRb} has been increasingly applied to improve code generation quality through visual rewards. VinciCoder~\cite{VinciCoder} uses DINOv2-based visual similarity as a reward for group relative policy optimization (GRPO)~\cite{GRPO}. VisRefiner~\cite{VisRefiner} designs a hierarchical reward including format, refinement, and quality components, also optimized with GRPO. While these methods demonstrate the effectiveness of visual reward signals, they suffer from crucial limitations. First, the reward is only provided at the end of each generation episode, leading to sparse feedback. Additionally, they lack a mechanism to assign credit to individual reflection and coding actions within a multi-step trajectory.
We address these limitations through a novel counterfactual-based reinforcement learning framework. By constructing two baselines - a ``no-reflection'' baseline to isolate the value of reflection and a ``given-reflection'' baseline to evaluate action quality - the method assigns dense, step-level advantage signals to both reflection and execution actions. This approach overcomes the reward sparsity problem that plagues standard policy gradient methods like PPO and GRPO when applied to multi-turn reflection trajectories.

\subsection{Self-Reflection and Iterative Refinement in LLMs}

The concept of self-reflection has been explored in various LLM-based reasoning tasks. ChartSketcher~\cite{ChartSketcher} introduces a Sketch-CoT mechanism where models annotate intermediate reasoning steps onto charts via a programmatic sketching library, iteratively feeding visual annotations back into reasoning. However, its reflection is grounded in visual sketches rather than code execution outputs, and it does not enforce consistency between reflection and subsequent actions. TinyChart~\cite{TinyChart} employs Program-of-Thoughts (PoT) learning for numerical reasoning but lacks iterative refinement.
Empirical studies on reflection effectiveness reveal two major challenges: models often generate superficial or incorrect reflections (ineffective reflection), and even when reflections are accurate, models frequently fail to implement the suggested changes in the subsequent code (inconsistent execution). Our proposed MCTS sampling with consistency pruning and multi-step reinforcement learning ensures that every output trajectory exhibits strictly increasing rewards and strict adherence to preceding reflections.





\newtcolorbox{promptbox}[1]{
    enhanced,
    colback=white,                  
    colframe=black,                 
    coltitle=white,                 
    fonttitle=\bfseries\small,      
    fontupper=\small,               
    attach boxed title to top left={
        yshift=-2mm, 
        xshift=0mm
    },
    boxed title style={
        sharp corners=downhill,     
        colback=black,              
        colframe=black,
        arc=3pt,                    
    },
    title=#1,                       
    arc=4pt,                        
    boxrule=1pt,                    
    top=12pt,                       
    bottom=8pt,
    left=8pt,
    right=8pt
}


\begin{figure*}[t]
\begin{promptbox}{System Prompt}
You are an expert in reproducing scientific figures and charts. Given a chart image, you generate a Python script to reproduce it accurately. After thorough reasoning, strictly output code in this format:

\begin{verbatim}
<code>
```python
import matplotlib
matplotlib.use('Agg')
import matplotlib.pyplot as plt
import numpy as np
import pandas as pd
import seaborn as sns

(Your generated code goes here)

plt.tight_layout()
plt.savefig("output.png", dpi=150, bbox_inches="tight")
```
</code>
\end{verbatim}
\end{promptbox}
\end{figure*}




\begin{figure*}[t] 
\begin{promptbox}{Environmental Feedback (Debug)}
Code execution error, error message is: \\[4pt]
\texttt{[Execution Status, Error Traces]} \\[4pt]
Please analyze the cause of the error and try again.
\end{promptbox}
\end{figure*}

\begin{figure*}[t]
\begin{promptbox}{Environmental Feedback (Reflection)}
Please compare the differences between the generated image and the original reference image, summarize improvement methods, and output in the following format: \\[4pt]
<improvement> [suggestions for improvement] </improvement> \\[4pt]
<code> [improved code] </code> \\[4pt]
Based on the above analysis, if you believe that there is no need to make further modifications to the current code, simply output <improvement> DONE </improvement> and end the answer.
\end{promptbox}
\end{figure*}





\section{Benchmarks}
\label{sec:appendix_bench}
We evaluate our framework across three challenging benchmarks representing diverse layout granularities:
\begin{itemize}[leftmargin=*]
\item \textbf{RealChart2Code} is a complex, large-scale benchmark designed to evaluate code generation from real-world datasets across 50 plot types and multi-panel layouts~\cite{RealChart2Code}.
\item \textbf{ChartMimic} focuses on reproducing intricate academic-style charts from raw images, specifically testing the model's visual fidelity and layout structure through the \textit{Direct Mimic} and \textit{Customized Mimic} tasks~\cite{shi2024chartmimic}.
\item \textbf{Plot2Code} is composed of diverse visualization tasks mapping scientific paper figures directly to execution-ready code~\cite{wu2024plot2code}.
\end{itemize}

\subsection{Evaluation Metrics}
Functional correctness is measured via the \textbf{Pass Rate (\%)} (the ratio of generated code that compiles and runs without sandbox runtime errors)~\cite{RealChart2Code, wu2024plot2code}. To assess layout, typographic, and stylistic fidelity, we report the \textbf{Score} (or \textbf{Rating}) evaluated by standard multimodal models-as-a-judge (such as GPT-4o~\cite{gpt4o})~\cite{shi2024chartmimic}. For code-level comparisons, we report token-level \textbf{Text} similarity~\cite{wu2024plot2code}. 

\section{Dataset}
\label{sec:appendix_dataset}

\subsection{Data Collection and Filtering Pipeline}
\label{subsec:data_pipeline}
To cultivate a robust understanding of academic-grade plotting structures and diverse scientific styles, we curate a high-quality visualization training set extracted directly from contemporary computer science and AI research publications.

\paragraph{Source Document Collection.} We gather a corpus of 50K PDF papers published in the past three years. To ensure high graphic standards and rigorous formatting, we target publications from top-tier venues (AAAI, ACL, NeurIPS, ICLR, EMNLP, ICML, NAACL) alongside raw preprints from arXiv.

\paragraph{Figure Extraction.} For all collected PDF manuscripts, we extract embedded figures and spatial coordinates using MinerU~\cite{mineru}, an open-source document parsing toolkit capable of isolating structured visual elements from multi-column scientific layouts.

\paragraph{Figure Filtering.} Scientific papers contain diverse visual structures (e.g., system diagrams, photos) unsuitable for chart replication code learning. We use VisJudge~\cite{visjudge}, an automated aesthetic and quality assessment model, to filter out non-chart images, heavily distorted plots, structure-naive figures, and low-resolution graphics.

\paragraph{Dataset Derivation.} Following filtering, we obtain \textbf{over 60K high-quality chart images}, spanning both standalone plots and complex multi-panel subplots. We allocate 50K instances for the Stage-1 warm-up SFT, and retain the remaining 10K instances as training references for Stage-2 reinforcement learning.

\subsection{Train-Test De-duplication Protocol}
\label{subsec:deduplication}
To ensure our evaluation measures true generalization rather than rote memorization, we clarify the benchmark sources and enforce a strict de-duplication pipeline prior to training.

We computed visual similarity between our candidate 60K training pool and all benchmark test sets using three progressive criteria:
\begin{enumerate}[leftmargin=*]
    \item \textbf{Exact Matches:} Cryptographic pixel hashing (MD5) to eliminate identical image files.
    \item \textbf{Perceptual Hashing (pHash):} Identification of near-duplicates differing only in resolution, compression, or margin padding. We enforce a strict Hamming distance threshold of $\le 4$, a conservative standard in image deduplication engines that isolates identical visual targets with near-zero false matches.
    \item \textbf{CLIP-based Semantic Filtering:} Identification of charts with highly similar layouts and visual representations. We employ a Cosine Similarity threshold of $\ge 0.88$ using CLIP \texttt{ViT-L/14}~\cite{clip} embeddings, following standard web-scale semantic pruning practices~\cite{Abbas2023SemDeDup,nemo_curator}. For ChartMimic charts with available metadata, we also cross-match arXiv paper identifiers and exclude figures originating from identical source manuscripts.
\end{enumerate}

Any training instance flagged by these criteria was permanently removed prior to training. Table~\ref{tab:deduplication_stats} presents the detailed filtering statistics.

\begin{table*}[!htbp]
\centering
\small
\caption{Statistics of the multi-stage sample-level de-duplication pipeline across evaluation benchmarks.}
\label{tab:deduplication_stats}
\resizebox{\linewidth}{!}{
\begin{tabular}{lcccc}
\toprule
\textbf{Benchmark} & \textbf{Exact Match} & \textbf{pHash} & \textbf{CLIP Threshold} & \textbf{Filtered from Train (\%)} \\
\midrule
\textbf{RealChart2Code}~\cite{RealChart2Code} & 0 (0.00\%) & Hamming $\le 4$ & Cosine sim. $\ge 0.88$ & 0 (0.00\%) \\
\textbf{ChartMimic}~\cite{shi2024chartmimic}     & 0 (0.00\%) & Hamming $\le 4$ & Cosine sim. $\ge 0.88$ & 154 (0.26\%) \\
\textbf{Plot2Code}~\cite{wu2024plot2code}      & 0 (0.00\%) & Hamming $\le 4$ & Cosine sim. $\ge 0.88$ & 50 (0.08\%) \\
\bottomrule
\end{tabular}
}
\end{table*}

\subsection{Latent Space Distribution Analysis}
\label{subsec:latent_space}
To verify the diversity and coverage of our curated training set ($N \approx 60,000$), we analyze its semantic representation against the evaluation benchmarks. We use the vision encoder of CLIP (\texttt{ViT-L/14}) to extract 768-dimensional visual feature vectors, subsequently projected to a 2D manifold using UMAP~\cite{umap} with a Cosine Distance metric.

\begin{figure}[!t]
\centering
\includegraphics[width=1.0\linewidth]{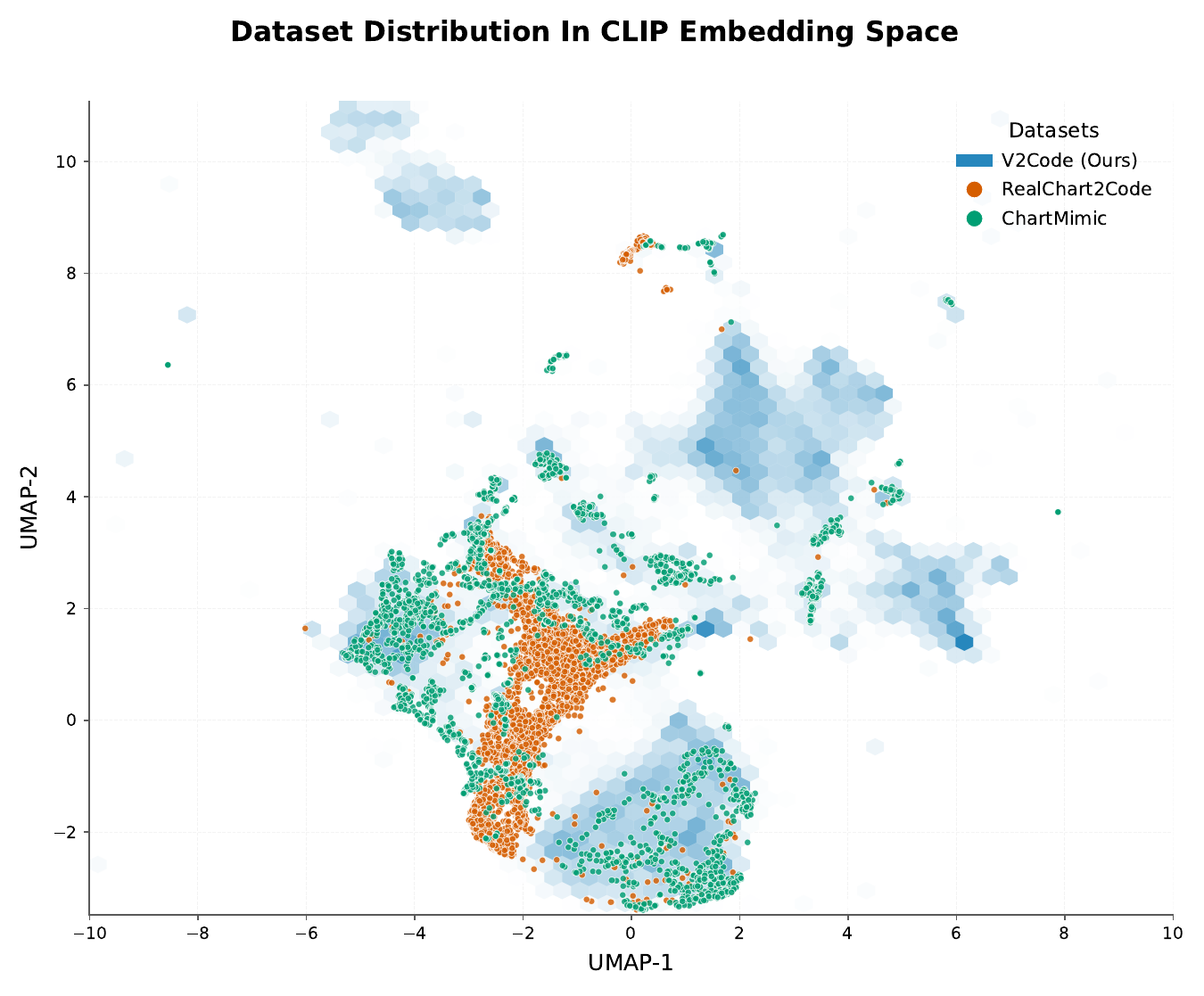}
\caption{2D manifold projection of visual representations across datasets.}
\label{fig:manifold}
\end{figure}

As illustrated in Figure~\ref{fig:manifold}, our training dataset establishes a broad, continuous latent foundation. The distributions of ChartMimic~\cite{shi2024chartmimic} and RealChart2Code~\cite{RealChart2Code} exhibit substantial distributional overlap with our corpus without forming disjoint clusters. This confirms strong domain alignment, while the multi-stage filtering in $\S$\ref{subsec:deduplication} guarantees sample-level non-contamination.

\section{Implementation Details}
\label{sec:appendix_training_and_hyperparams}

\subsection{Sandboxed Sandbox Environment}
\label{sec:appendix_env}
To prevent code execution safety hazards and standardize environmental variables, all code snippets generated during evaluation and MCTS rollout are executed inside isolated Docker containers based on Python 3.13. Pre-installed libraries include \texttt{pandas (2.2)}, \texttt{numpy (2.1)}, \texttt{matplotlib (3.9.2)}, \texttt{seaborn (0.13.2)}, and \texttt{scipy (1.14)}. Each container is restricted to 1 CPU core and 4 GB RAM, with a strict execution timeout of 120 seconds.

\subsection{Training Configuration}
\label{subsec:model_config}
We instantiate ViCo using two open-source vision-language backbones: \textbf{Qwen3-VL-4B} and \textbf{Qwen3-VL-8B}~\cite{Qwen3VL}. The 4B model serves as a compute-efficient agent, while the 8B variant leverages expanded multimodal capacity to model intricate spatial layouts. 

During training, both the visual projection connector and the language model backbone are unfrozen for end-to-end multimodal alignment, while the visual encoder is kept frozen to stabilize visual feature extraction and minimize peak memory usage. All training sessions are accelerated with DeepSpeed ZeRO-3~\cite{ZeRO} on an 8$\times$NVIDIA A800 GPU node. Table~\ref{tab:train_config} reports the exact parameter setup distinguished by training phases.

\subsection{Hyperparameter Settings}
\label{subsec:hyperparameters}
To permanently resolve notation overloading and provide a unified reference, we consolidate all hyperparameters across the MCTS warm-up stage, multi-step PPO reinforcement learning, and the HHLG evaluation framework into Table~\ref{tab:unified_hyperparams}. 

\begin{table}[!tbp]
\centering
\small
\caption{Hyperparameter settings for Phase 1 (Self-Supervised Fine-tuning, SSFT) and Phase 2 (Multi-step Reinforcement Learning, MSRL) training of ViCo.}
\label{tab:train_config}
\resizebox{\linewidth}{!}{
\begin{tabular}{lcc}
\toprule
\textbf{Hyperparameter} & \textbf{Phase 1 (SSFT)} & \textbf{Phase 2 (MSRL)} \\
\midrule
Backbone Model & \multicolumn{2}{c}{Qwen3-VL~\cite{Qwen3VL}} \\
Optimizer & \multicolumn{2}{c}{AdamW~\cite{adamw}} \\
Learning Rate & $1\times 10^{-5}$ & $5\times 10^{-6}$ \\
Learning Rate Schedule & Cosine decay & Cosine decay \\
Warmup Ratio & 0.03 & 0.05 \\
Batch Size & 128 & 64 \\
Discount Factor ($\gamma$) & -- & 0.95 \\
Counterfactual Samples ($K$) & -- & 5 \\
Max Sequence Length (Tokens) & 32,768 & 32,768 \\
\bottomrule
\end{tabular}
}
\end{table}

\begin{table*}[!htbp]
\centering
\scriptsize
\caption{Comprehensive hyperparameter settings across the MCTS warm-up stage, multi-step PPO reinforcement learning, and the HHLG-based reward evaluation framework. Overloaded notations from prior versions have been systematically disambiguated.}
\label{tab:unified_hyperparams}
\resizebox{\linewidth}{!}{
\begin{tabular}{lccl}
\toprule
\textbf{Hyperparameter} & \textbf{Symbol} & \textbf{Value} & \textbf{Explanation} \\
\midrule
\multicolumn{4}{l}{\textit{\textbf{Stage 1: MCTS-Based Trajectory Synthesis \& Warm-Up}}} \\
\midrule
Max reflection steps    & $K$              & 3     & Maximum iteration depth of the multi-turn coding loop \\
Branching factor        & $C$              & 8     & Number of candidate code expansions generated per node \\
Exploration constant    & $c$              & 1.25  & Standard PUCT exploration weight \\
Maximum patience        & $\rho_{\max}$    & 2     & Sequence tolerance for consecutive non-improving steps \\
Penalty coefficient     & $\beta$          & 0.8   & Scaling factor for differential history penalty \\
Consistency threshold   & $\tau$           & 0.8   & Strictness limit for matching reflection text to code block \\
Consistency gain        & $\lambda$        & 0.2   & Weight coefficient of consistency reward component \\
\midrule
\multicolumn{4}{l}{\textit{\textbf{Stage 2: Multi-Step PPO Reinforcement Learning}}} \\
\midrule
Counterfactual samples  & $M$              & 8     & Parallel rollouts used to compute step-wise advantage baselines \\
Discount factor         & $\gamma_{rl}$    & 1.0   & Temporal decay rate for downstream credit assignment \\
Clip range              & $\epsilon$       & 0.2   & PPO surrogate objective clipping boundary \\
\midrule
\midrule
\multicolumn{4}{l}{\textit{\textbf{Reward Engine: HHLG-Based Multi-Dimensional Perceptual Reward}}} \\
\midrule
Semantic weight         & $w_1$            & 0.25  & Importance weight of text OCR alignment component ($S_{\text{sem}}$) \\
Layout weight           & $w_2$            & 0.25  & Importance weight of macro/micro graph layout component ($S_{\text{lay}}$) \\
Style weight            & $w_3$            & 0.25  & Importance weight of VGG Gram texture component ($S_{\text{style}}$) \\
Color weight            & $w_4$            & 0.25  & Importance weight of CIELAB visual mask color component ($S_{\text{color}}$) \\
Bipartite cost weight   & $\alpha$         & 0.5   & Balances Levenshtein distance vs. IoU in semantic bipartite matching \\
Layout scale factor     & $\beta_{lay}$    & 0.1   & Scaling factor mapping GED graph distances to layout scores \\
Style scale factor      & $\gamma_{st}$    & 0.05  & Scaling factor mapping Gram matrix distances to style scores \\
\midrule
Matched visual pairs    & $\mathcal{P}$    & --    & Index set of matched visual node pairs from layout alignment \\
\bottomrule
\end{tabular}
}
\end{table*}

\section{Theoretical and Empirical Analysis of Reflection Advantage}
\label{sec:appendix_reflection_analysis}

\subsection{Logical Derivation: Quality vs. Presence}
\label{subsec:logical_proof}
A critical theoretical question is whether the reflection advantage $A(r_t) = Q_{\rm mid}(h_t, r_t) - Q_{\rm mid}(h_t, r_\emptyset)$ merely rewards the superficial \textit{presence} of critique tokens, or genuinely measures the \textit{content quality and utility} of the specific critique $r_t$.

Because downstream coding actions $a_t$ are sampled directly from the conditioned policy $\pi_\theta(a_t \mid h_t, r_t)$, the expected intermediate value $Q_{\rm mid}(h_t, r_t)$ causally inherits the validity of $r_t$:
\begin{itemize}[leftmargin=*]
    \item \textbf{Case 1: Misleading Critique (Poor Content):} If $r_t$ contains hallucinated or erroneous diagnoses (e.g., misclassifying plot types or prescribing incorrect coordinate transformations), it actively biases the subsequent coding policy. The downstream execution is likely to fail or introduce severe graphical deviations. Consequently, the expected reward under this flawed reflection is lower than executing directly without critique ($r_\emptyset$):
    \[Q_{\rm mid}(h_t, r_t) < Q_{\rm mid}(h_t, r_\emptyset) \implies A(r_t) < 0.\]
    The policy is thus explicitly penalized for generating misleading reflections despite their presence.
    \item \textbf{Case 2: Constructive Critique (High Content Quality):} If $r_t$ correctly diagnoses visual discrepancies and provides actionable instructions, it guides the subsequent coding policy to resolve rendering errors:
    \[Q_{\rm mid}(h_t, r_t) > Q_{\rm mid}(h_t, r_\emptyset) \implies A(r_t) > 0.\]
    The policy receives positive advantage because the critique yielded positive marginal utility.
\end{itemize}
Therefore, $r_\emptyset$ functions as an unbiased, neutral control baseline, allowing $A(r_t)$ to measure the true marginal contribution of the reflection's content.

\subsection{Computational Tractability of Alternative Baselines}
\label{subsec:computational_bottleneck}
The theoretical ideal for isolating reflection content is a fully marginalized content-contrastive baseline:
\[
A_{\rm contrast}(r_t) = Q_{\rm mid}(h_t, r_t) - \frac{1}{J} \sum_{j=1}^J Q_{\rm mid}(h_t, r'_j),
\]
where $\{r'_1, \dots, r'_J\}$ are alternative candidate reflections sampled from $\pi_\theta(\cdot \mid h_t)$.

However, computing $A_{\rm contrast}$ is computationally prohibitive. Because rendered code must be compiled and executed inside isolated Docker containers to obtain visual rewards, evaluating $Q_{\rm mid}$ across $J$ reflections requires sampling $M$ coding rollouts each, demanding $\mathcal{O}(J \times M)$ sandbox compilation cycles per optimization step. With $J=8$ and $M=8$, this scales training time from tens of GPU hours to hundreds of hours. By contrast, our neutral baseline $r_\emptyset$ requires only $\mathcal{O}(2M)$ rollouts, ensuring stable advantage estimation within accessible academic compute limits.

\subsection{Empirical Verification of Advantage Correlation}
\label{subsec:advantage_correlation}
To empirically verify that $A(r_t)$ aligns with actual step-level visual improvements, we conduct a post-hoc correlation analysis over 200 held-out validation trajectories (comprising 600 discrete reflection-coding refinement steps). For each step $t$, we measure:
\begin{enumerate}[leftmargin=*]
    \item The estimated reflection advantage $A(r_t)$.
    \item The ground-truth visual consistency gain: $\Delta R_{\rm true} = R(t) - R(t-1)$, where $R(t)$ denotes the HHLG visual fidelity score of the rendered chart at step $t$.
\end{enumerate}

Statistical analysis reveals a strong, statistically significant positive correlation between the estimated advantage and the true visual gain:
\begin{itemize}[leftmargin=*]
    \item \textbf{Spearman Rank Correlation ($\rho$):} $\mathbf{0.742}$ ($p < 0.001$)
    \item \textbf{Pearson Correlation ($r$):} $\mathbf{0.718}$ ($p < 0.001$)
\end{itemize}
This tight alignment confirms that $A(r_t)$ reliably captures meaningful improvements in visual fidelity rather than rewarding token presence.

\section{More Ablations Results}
\label{sec:appendix_supp_exp}

\subsection{Stage-Level Performance Decomposition}
\label{subsec:stage_ablation}
Because our training corpus contains raw rasterized chart images without ground-truth plotting scripts or human reflection logs, constructing a 50K human-written baseline is practically infeasible. To decouple the gains from in-domain supervision, MCTS trajectory curation, and multi-step RL, we construct a strong pseudo-labeled baseline using Qwen3-VL-235B (Teacher). Given an initial program, its rendered image, sandbox execution logs, and the target image, the teacher generates a single fixed reflection-code pair without branching, consistency checks, or MCTS pruning.

\begin{table}[!t]
\centering
\small
\caption{Stage-level performance decomposition of ViCo (8B) evaluated on RealChart2Code and ChartMimic (Direct Mimic).}
\label{tab:stage_ablation}
\resizebox{\columnwidth}{!}{
\begin{tabular}{l cccc}
\toprule
\multirow{2}{*}{\textbf{Training Configuration}} & \multicolumn{2}{c}{\textbf{RealChart2Code}} & \multicolumn{2}{c}{\textbf{ChartMimic}} \\
\cmidrule(lr){2-3} \cmidrule(lr){4-5}
 & Pass (\%) & Score & Pass (\%) & Score \\
\midrule
Vanilla SFT (Teacher Pseudo-labels) & 85.2 & 2.4 & 90.5 & 68.4 \\
\midrule
\rowcolor{gray!10} \multicolumn{5}{l}{\textit{Single-Stage Control Baselines}} \\[4pt]
\quad MCTS Warm-up (w/o RL)              & 89.4 & 2.6 & 92.1 & 72.8 \\
\quad Cold-start RL (w/o Warm-up)        & 62.5 & 1.8 & 71.3 & 45.2 \\
\midrule
\rowcolor{gray!10} \textbf{Full ViCo (Warm-up + RL)} & \textbf{98.8} & \textbf{3.4} & \textbf{99.2} & \textbf{81.5} \\
\bottomrule
\end{tabular}
}
\end{table}

As shown in Table~\ref{tab:stage_ablation}, MCTS warm-up outperforms static teacher SFT by enforcing reflection-execution consistency. Furthermore, while cold-start RL struggles due to the high initial search entropy of multi-turn code generation, pairing MCTS warm-up with counterfactual RL yields substantial cumulative gains across both compilation pass rates and visual scores.

\subsection{Leave-One-Component-Out Reward Ablation}
\label{subsec:loo_ablation}
To isolate the marginal contribution of each perceptual dimension within our HHLG reward framework, we conduct a Leave-One-Component-Out (LOO) ablation on the ViCo-4B policy. We re-train both the SFT warm-up and RL stages under four ablated configurations by setting each component weight to $0.0$ in turn (with the remaining three dimensions weighted equally at $0.33$). Table~\ref{tab:loo_ablation} summarizes the execution pass rates, overall benchmark visual scores, and all four individual HHLG sub-scores on ChartMimic.

\begin{table}[!htbp]
\centering
\scriptsize
\caption{Leave-One-Component-Out (LOO) ablation of the HHLG reward formulation across SFT warm-up and RL stages using ViCo-4B evaluated on ChartMimic.}
\label{tab:loo_ablation}
\resizebox{\linewidth}{!}{
\begin{tabular}{l cccccc}
\toprule
\textbf{Training Configuration} & \textbf{Pass (\%)} & \textbf{Score} & $\mathbf{S_{\rm sem}}$ & $\mathbf{S_{\rm lay}}$ & $\mathbf{S_{\rm style}}$ & $\mathbf{S_{\rm color}}$ \\
\midrule
\multicolumn{7}{l}{\textit{\textbf{Phase 1: SFT Warm-Up Stage}}} \\
\midrule
\rowcolor{gray!10} \textbf{ViCo-4B (Warmup-only)} & \textbf{89.1} & \textbf{69.4} & \textbf{75.4} & \textbf{69.5} & \textbf{63.8} & \textbf{65.2} \\
\quad w/o Semantic & 88.2 & 44.8 & 32.2 & 70.8 & 64.3 & 65.6 \\
\quad w/o Layout & 88.3 & 46.5 & 74.8 & 38.6 & 58.1 & 60.2 \\
\quad w/o Style & 88.5 & 58.2 & 75.6 & 69.2 & 36.4 & 64.9 \\
\quad w/o Color & 88.7 & 57.3 & 75.1 & 69.4 & 63.4 & 34.8 \\
\midrule
\multicolumn{7}{l}{\textit{\textbf{Phase 2: Multi-Step RL Stage}}} \\
\midrule
\rowcolor{gray!10} \textbf{ViCo-4B (Full Framework)} & \textbf{98.1} & \textbf{78.8} & \textbf{84.5} & \textbf{78.8} & \textbf{72.6} & \textbf{74.3} \\
\quad w/o Semantic & 97.2 & 51.5 & 38.2 & 80.5 & 73.1 & 74.9 \\
\quad w/o Layout & \textbf{98.1} & 53.4 & 83.9 & 42.4 & 66.8 & 68.3 \\
\quad w/o Style & 97.6 & 68.2 & 84.7 & 78.4 & 44.5 & 73.8 \\
\quad w/o Color & 97.9 & 66.4 & 84.2 & 78.6 & 72.2 & 40.5 \\
\bottomrule
\end{tabular}
}
\end{table}

The empirical breakdown confirms three key properties:
\begin{enumerate}[leftmargin=*]
    \item \textbf{Decoupled Sensitivity:} Disabling any single sub-reward triggers a sharp collapse in its corresponding sub-score (e.g., $S_{\rm sem}$ drops from $84.5$ to $38.2$ when $w_{\rm sem}=0$), while unaffected dimensions remain stable or experience minor relaxation gains due to reduced optimization constraints.
    \item \textbf{Necessity of Full Composition:} The composite visual score experiences severe degradation across all ablated settings in both stages, proving that visual chart replication requires balanced multi-objective constraints to avoid degenerate solutions (e.g., generating correct layout geometry with mismatched semantic labels).
\end{enumerate}

\subsection{Extended RL Training Horizon and Reward Hacking Analysis}
\label{subsec:extended_training}
To examine whether HHLG is susceptible to delayed Goodhart's law effects or reward hacking under extended policy updates, we prolong policy optimization to $2\times$ the standard training horizon (up to 2,000 steps). We periodically benchmark checkpoints on RealChart2Code by evaluating internal HHLG reward alongside independent GPT-4o visual judging scores.

\begin{figure}[!t]
\centering
\includegraphics[width=\linewidth]{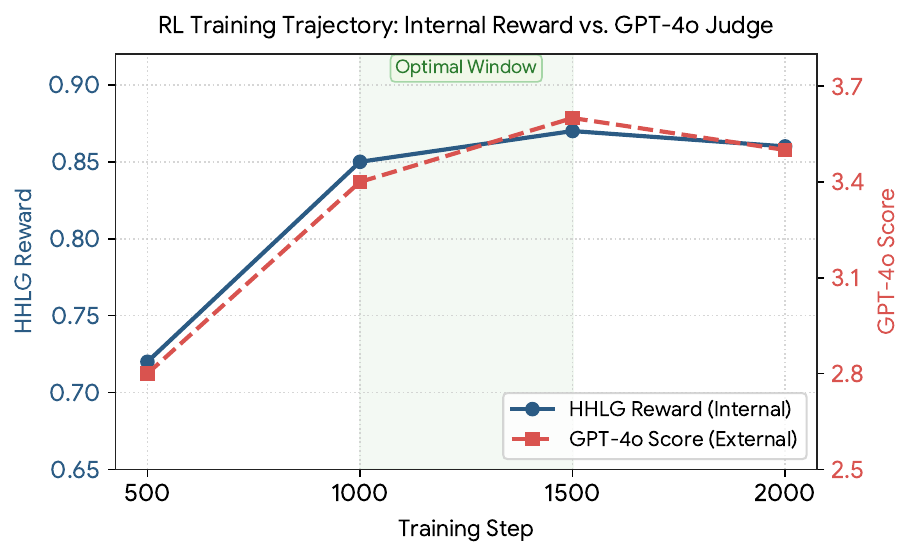}
\caption{Trajectory of internal HHLG rewards versus independent GPT-4o evaluation scores under extended training horizons on RealChart2Code. Both metrics align synchronously, demonstrating no reward hacking up to 2,000 steps.}
\label{fig:extended_training}
\end{figure}

As demonstrated in Figure~\ref{fig:extended_training}, the internal HHLG reward and independent GPT-4o evaluations scale synchronously without divergent divergence up to 2,000 steps. Beyond step 1,500, visual fidelity exhibits mild plateauing ($3.6 \to 3.5$) without catastrophic degradation. This highlights the structural robustness of discrete graph-grounded rewards over unconstrained neural judges, confirming that early stopping near 1,000--1,500 steps is optimal for computational efficiency.

\section{More Analyses}

\subsection{Parsing Sensitivity and Error Accumulation}
\label{subsec:sensitivity_analysis}
Because HHLG computes semantic, layout, style, and color metrics in parallel from a shared graph, an error in one metric does not structurally propagate into others. However, upstream parsing noise (e.g., OCR or segmentation failures) could affect common downstream node extractors. We evaluate this robustness by injecting synthetic noise into 200 human-annotated examples and measuring metric perturbations alongside shifts in Spearman correlation with human judgments ($\Delta\rho_{\rm human}$).

\begin{table*}[!htbp]
\centering
\small
\caption{Sensitivity analysis of HHLG sub-metrics and human judgment correlation ($\Delta\rho_{\rm human}$) under controlled upstream parsing noise.}
\label{tab:sensitivity_analysis}
\begin{tabular}{l c c c}
\toprule
\textbf{Parsing Perturbation} & \textbf{Extent} & \textbf{Sub-metric Impact ($\Delta S_{\rm sem} / \Delta S_{\rm lay} / \Delta S_{\rm style} / \Delta S_{\rm color}$)} & $\mathbf{\Delta\rho_{\rm human}}$ \\
\midrule
OCR token corruption     & 10\% & ($-0.10$, $-0.02$, $\phantom{-}0.00$, $\phantom{-}0.00$) & $-0.015$ \\
Visual-node dropout      & 10\% & ($\phantom{-}0.00$, $-0.08$, $-0.01$, $-0.07$)          & $-0.035$ \\
Mask erosion/dilation    & 10\% & ($\phantom{-}0.00$, $-0.01$, $\phantom{-}0.00$, $-0.04$) & $-0.012$ \\
Text-box jitter          & 5\%  & ($\phantom{-}0.00$, $-0.03$, $\phantom{-}0.00$, $\phantom{-}0.00$) & $-0.008$ \\
Subplot-boundary jitter  & 5\%  & ($-0.02$, $-0.06$, $-0.05$, $-0.03$)                     & $-0.042$ \\
\bottomrule
\end{tabular}
\end{table*}

Table~\ref{tab:sensitivity_analysis} demonstrates that local parsing errors remain strictly confined within their respective modules (e.g., OCR corruption affects only $S_{\rm sem}$). While macro-level partition shifts (subplot jitter) exert a broader influence, the overall human correlation degradation remains minimal ($\Delta\rho_{\rm human} = -0.042$), confirming that the composite metric is highly resilient to parsing imperfections.

\subsection{Controlled-Budget Inference and Fairness Evaluation}
\label{subsec:budget_fairness}
To ensure a strictly fair comparison with respect to inference-time compute, we evaluate models on ChartMimic under uniform execution budgets, restricting the refinement loop to 0 (single-shot / Pass@1), 1, and 2 reflection rounds.

\begin{table}[!t]
\centering
\small
\caption{Code execution pass rates (\%) under strictly controlled inference budgets on ChartMimic.}
\label{tab:budget_fairness}
\resizebox{\columnwidth}{!}{
\begin{tabular}{l ccc}
\toprule
\textbf{Model} & \textbf{Pass@1 (Single-shot)} & \textbf{Reflect$\times$1} & \textbf{Reflect$\times$2} \\
\midrule
Qwen3-VL-8B       & 65.6          & 63.4          & 65.2 \\
Qwen3-VL-30B      & 89.5          & 90.5          & 90.8 \\
Qwen3-VL-235B     & \textbf{95.5} & \textbf{98.3} & \textbf{99.5} \\
ChartCoder        & 90.9          & 83.2          & 79.5 \\
\midrule
\rowcolor{gray!10} \textbf{ViCo-8B} & 88.5 & 96.7 & 99.2 \\
\bottomrule
\end{tabular}
}
\end{table}

As presented in Table~\ref{tab:budget_fairness}:
\begin{enumerate}[leftmargin=*]
    \item \textbf{Single-Shot Strength:} In single-shot generation (Pass@1), ViCo-8B achieves an $88.5\%$ pass rate, significantly outperforming base Qwen3-VL-8B ($65.6\%$) and nearing the 30B model ($89.5\%$), showing that reflective warm-up enhances base code synthesis.
    \item \textbf{Monotonic Multi-Turn Scaling:} While uncalibrated baselines (such as ChartCoder) suffer from regressive degradation across multiple edits ($90.9\% \to 79.5\%$), ViCo systematically resolves runtime errors over successive rounds, scaling to $99.2\%$ at Reflect$\times$2 and matching the 235B model.
\end{enumerate}

\subsection{Comparison with Trajectory-Level RL Baselines}
\label{subsec:grpo_comparison}
To validate the effectiveness of our step-wise counterfactual credit assignment against holistic trajectory-level optimization, we compare ViCo MSRL against trajectory-level PPO and GRPO baselines. All methods are initialized from the identical MCTS warm-up checkpoint and trained with equal compute budgets and HHLG reward feedback.

\begin{table}[!t]
\centering
\small
\caption{Performance comparison between holistic trajectory-level RL baselines and ViCo multi-step counterfactual RL (MSRL) on RealChart2Code and ChartMimic.}
\label{tab:grpo_comparison}
\resizebox{\columnwidth}{!}{
\begin{tabular}{l cccc}
\toprule
\multirow{2}{*}{\textbf{RL Method}} & \multicolumn{2}{c}{\textbf{RealChart2Code}} & \multicolumn{2}{c}{\textbf{ChartMimic}} \\
\cmidrule(lr){2-3} \cmidrule(lr){4-5}
 & Pass (\%) & Score & Pass (\%) & Score \\
\midrule
Warmup + GRPO (Terminal reward) & 49.7 & 3.0 & 89.2 & 69.7 \\
Warmup + PPO (Terminal reward)  & 57.1 & 3.0 & 89.4 & 77.1 \\
\midrule
\rowcolor{gray!10} \textbf{ViCo (Warmup + MSRL)} & \textbf{98.8} & \textbf{3.4} & \textbf{99.2} & \textbf{81.5} \\
\bottomrule
\end{tabular}
}
\end{table}

As shown in Table~\ref{tab:grpo_comparison}, standard trajectory-level GRPO and PPO struggle in alternating reflection-coding environments due to severe credit blurring over long horizons, achieving suboptimal pass rates (49.7\% and 57.1\% on RealChart2Code). In contrast, ViCo's counterfactual advantage decomposition provides dense, step-specific training signals, substantially boosting both compilation reliability and visual fidelity.

\subsection{Computational Cost, Memory, and Throughput Profiling}
\label{subsec:computational_cost}
We benchmark the training efficiency, throughput, and hardware footprints of ViCo against standard multi-turn RL baselines on an 8$\times$NVIDIA A800 GPU node across both 4B and 8B model scales in Tables~\ref{tab:profiling_4b} and \ref{tab:profiling_8b}.

\begin{table}[!t]
\centering
\small
\caption{Training resource profiling for the 4B parameter model on 8$\times$NVIDIA A800 GPUs.}
\label{tab:profiling_4b}
\resizebox{\columnwidth}{!}{
\begin{tabular}{lcccc}
\toprule
\textbf{Method} & \textbf{GPU Setup} & \textbf{GPU Hours (h)} & \textbf{Peak Mem. (GiB)} & \textbf{Throughput (/s)} \\
\midrule
Warmup SFT      & 8$\times$A800 & 17.5 & 554.6 & 5.94 samples \\
Multi-turn GRPO & 8$\times$A800 & 14.2 & 527.4 & 10.80 rollouts \\
Multi-turn PPO  & 8$\times$A800 & 17.2 & 551.2 & 1.42 samples \\
ViCo MSRL       & 8$\times$A800 & 24.4 & 523.4 & 0.84 samples \\
\bottomrule
\end{tabular}
}
\end{table}

\begin{table}[!t]
\centering
\small
\caption{Training resource profiling for the 8B parameter model on 8$\times$NVIDIA A800 GPUs.}
\label{tab:profiling_8b}
\resizebox{\columnwidth}{!}{
\begin{tabular}{lcccc}
\toprule
\textbf{Method} & \textbf{GPU Setup} & \textbf{GPU Hours (h)} & \textbf{Peak Mem. (GiB)} & \textbf{Throughput (/s)} \\
\midrule
Warmup SFT      & 8$\times$A800 & 18.0 & 571.6 & 5.80 samples \\
Multi-turn GRPO & 8$\times$A800 & 18.3 & 525.5 & 6.92 rollouts \\
Multi-turn PPO  & 8$\times$A800 & 18.8 & 573.5 & 1.39 samples \\
ViCo MSRL       & 8$\times$A800 & 25.6 & 535.7 & 0.79 samples \\
\bottomrule
\end{tabular}
}
\end{table}

Two key empirical findings emerge:
\begin{enumerate}[leftmargin=*]
    \item \textbf{Academic Resource Feasibility:} On a standard single 8$\times$A800 node, full two-stage training converges within 25.6 GPU hours for the 8B model, bypassing heavy proprietary judges and operating entirely within academic laboratory limits.
    \item \textbf{Favorable Cost-Benefit Tradeoff:} ViCo MSRL requires only an additional 6.8 GPU hours over vanilla multi-turn PPO for the 8B model. This modest computational increase delivers decisive gains in execution pass rates (+41.7\% on RealChart2Code) and visual fidelity.
\end{enumerate}

\newpage
\includepdf[pages=-]{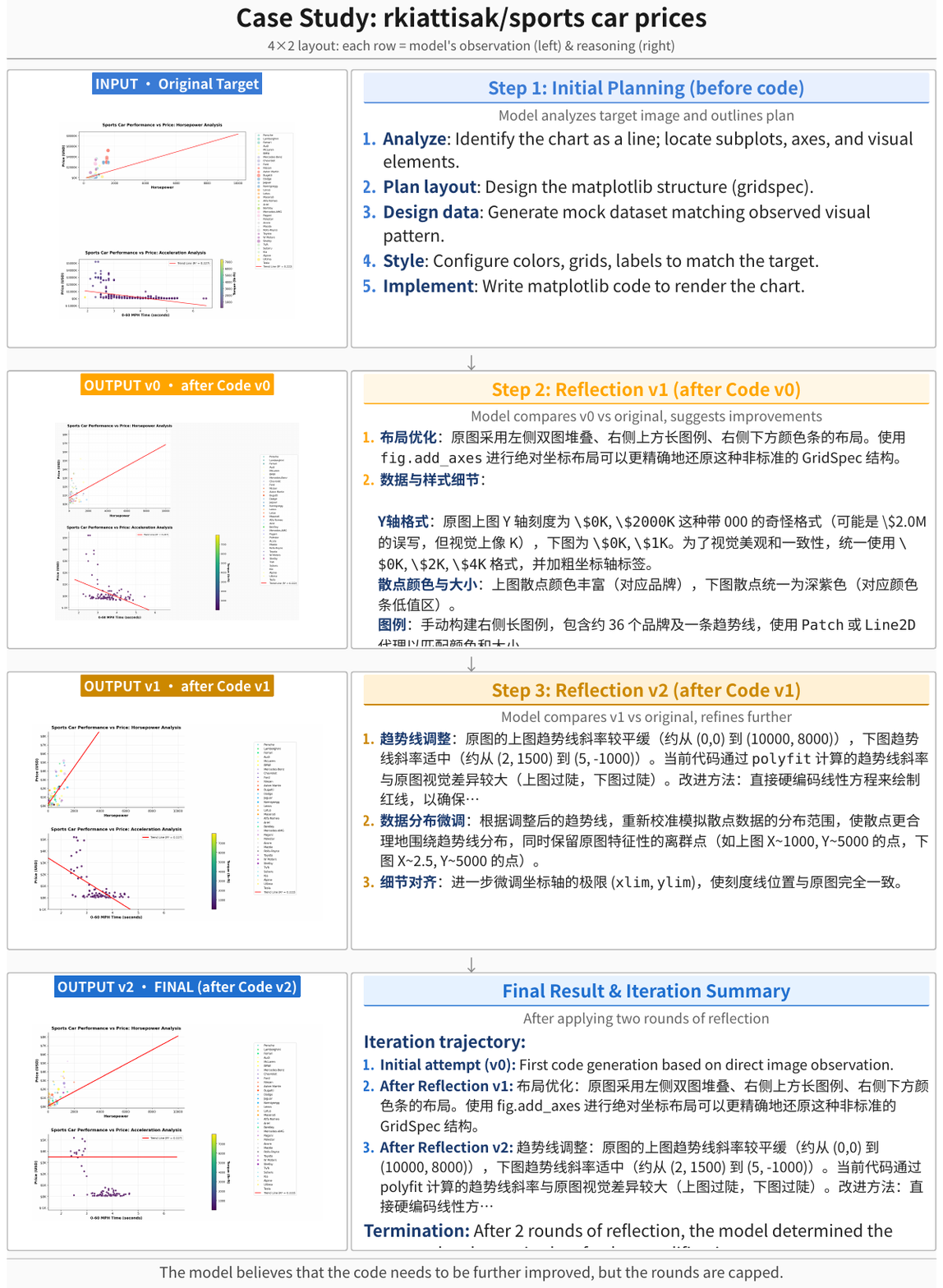}

\end{document}